\documentclass[letterpaper,10pt,conference]{ieeeconf}

\IEEEoverridecommandlockouts                           
\usepackage{graphics} % for pdf, bitmapped graphics files
\usepackage{epsfig} % for postscript graphics files
\usepackage{mathptmx} % assumes new font selection scheme installed
\usepackage{times} % assumes new font selection scheme installed
\usepackage{amsmath} % assumes amsmath package installed
\usepackage{amssymb}  % assumes amsmath package installed
\usepackage{multicol}
\usepackage[bookmarks=true]{hyperref}
\usepackage{graphicx}
\usepackage{tabularx}
\usepackage{cuted}
\usepackage{cite}
\usepackage{capt-of}
\usepackage{xcolor}
\usepackage{transparent}
\usepackage{colortbl}
\usepackage{amsmath} 
\usepackage{stfloats}
\usepackage{booktabs}
\usepackage[linesnumbered,ruled,vlined]{algorithm2e}
\usepackage{multirow}
\usepackage{makecell}
\usepackage[normalem]{ulem}
\usepackage{mathtools}

\makeatletter
\renewcommand{\@IEEEfigurecaptionsepspace}{\vskip 3pt\relax}
\renewcommand{\section}{\@startsection{section}{1}{\z@}{1.5ex}%
    {0.7ex}{\normalfont\normalsize\centering\scshape}}
\renewcommand{\subsection}{\@startsection{subsection}{2}{\z@}{1.5ex}%
    {0.7ex}{\normalfont\normalsize\itshape}}
\renewcommand{\subsubsection}{\@startsection{subsubsection}{3}{\parindent}{0ex}%
    {0ex}{\normalfont\normalsize\itshape}}
\renewcommand{\paragraph}{\@startsection{paragraph}{4}{2\parindent}{0ex}%
    {0ex}{\normalfont\normalsize\itshape}}
\makeatother
\definecolor{revisionblue}{RGB}{0,80,180}
\definecolor{tableheaderblue}{RGB}{214,224,241}

\title{\LARGE \bf
Any-ttach: Quick End-effector Swapping Enables Manipulation Dexterity with Simplicity
}

\author{
Weizhe Ni$^{*}$, Jinzhou Li$^{*}$, Xinyu Ma, Haoyu Li,  Eva Bonde, \\
Cody Andres Alessio-Bunnell, Wenjing Pan, Xianyi Cheng%
\thanks{$^{*}$These authors contributed equally to this work.}%
\thanks{All authors are with the Department of Mechanical Engineering and Materials Science, Duke University, Durham, NC 27708, USA.}
\thanks{\protect\rule{0pt}{12pt}%
This work has been submitted to the IEEE for possible publication.
Copyright may be transferred without notice, after which this
version may no longer be accessible.}
}

\begin{document}
\bstctlcite{IEEErefcontrol}

\maketitle
\pagestyle{empty}
% Let spare space fall at the column foot instead of stretching text gaps.
\raggedbottom

%%%%%%%%%%%%%%%%%%%%%%%%%%%%%%%%%%%%%%%%%%%%%%%%%%%%%%%%%%%%%%%%%%%%%%%%%%%%%%%%
\begin{abstract}
Robotic manipulation dexterity is often pursued by building increasingly complex high-DoF multifingered hands.
While many robotic hands are designed to replicate the morphology of human hands, the functional role of human hands suggests a different perspective: much of their complexity may exist to enable tool use and tool making.
This observation motivates \textit{Any-ttach}, a tool-centric manipulation system that leverages mechanical modularity and treats autonomous end-effector swapping as a pathway towards dexterity.
\textit{Any-ttach} combines: 1) a low-cost standardized interface with passive self-locking and mechanically triggered release mechanism; 2) a unified object-centric manipulation framework supporting kinesthetic teaching trajectory replay, human demonstration retargeting, and policy learning; and 3) hierarchical planning and execution system for tools and skills for multi-tool long-horizon tasks.
We demonstrated 15 heterogeneous end-effector modules supported by our system, spanning different robotic fingers, everyday tools (including articulated tools such as scissors), and unconventional end-effectors through the same interface.
Experiments demonstrate reliable autonomous swapping, automatic integration to new tools, and broad manipulation capability across different skill sources.
Two multi-tool application demonstrations, comprising six tool--skill steps across two tasks, further illustrate the autonomous composition of end-effector exchange and heterogeneous manipulation skills.
These results show that end-effector swapping can expand manipulation capability without concentrating increasing complexity entirely in a single end-effector.
More details and videos are available at \url{https://any-ttach.github.io/}.
\end{abstract}

%%%%%%%%%%%%%%%%%%%%%%%%%%%%%%%%%%%%%%%%%%%%%%%%%%%%%%%%%%%%%%%%%%%%%%%%%%%%%%%%
\section{Introduction}

General-purpose dexterity remains a central challenge in robotic manipulation. Most existing approaches treat dexterity as an intrinsic property of the end effector, seeking broader capability through high-degree-of-freedom(DoF) multifingered hands and complex hand--object contact interactions~\cite{qi2023hand,chen2024object,wang2022dexgraspnet,shaw2023leap}. 
% Although these systems demonstrate impressive performance, they tightly couple manipulation dexterity with hardware complexity and intricate contact-rich control.
Despite their mechanical and control complexity, many high-DoF hands are still primarily used for grasping and in-hand manipulation, with relatively limited demonstrations of diverse tool-use tasks. Extending a single hand to such tasks requires adapting to different tool geometries and maintaining stable grasps under task-specific contact forces, increasing the challenge of controlling a high-DoF mechanism and precise regulation of hybrid contact dynamics. These challenges become more clear in long-horizon tasks, where robots must coordinate low-level contact-rich manipulation with high-level skill sequencing.
% Although these systems enable grasping and in-hand manipulation, they rely on mechanically complex structure and complicated contact control.
% A single hand must accommodate many task-specific contact geometries, which increases execution complexity through coordination and control.
%% Even simple tasks require precise regulation of high dimensional contact dynamics, and the limitation becomes especially clear in long horizon tasks where robots must handle both low-level contact control and high-level skill sequencing.
% Even simple tasks require careful control of a high-DoF mechanism and precise regulation of hybrid dynamics of contacts, and the limitation becomes especially clear in long horizon tasks where robots must handle both low-level contact control and high-level skill sequencing.

\begin{figure}
    \centering
    \includegraphics[width=0.98\linewidth]{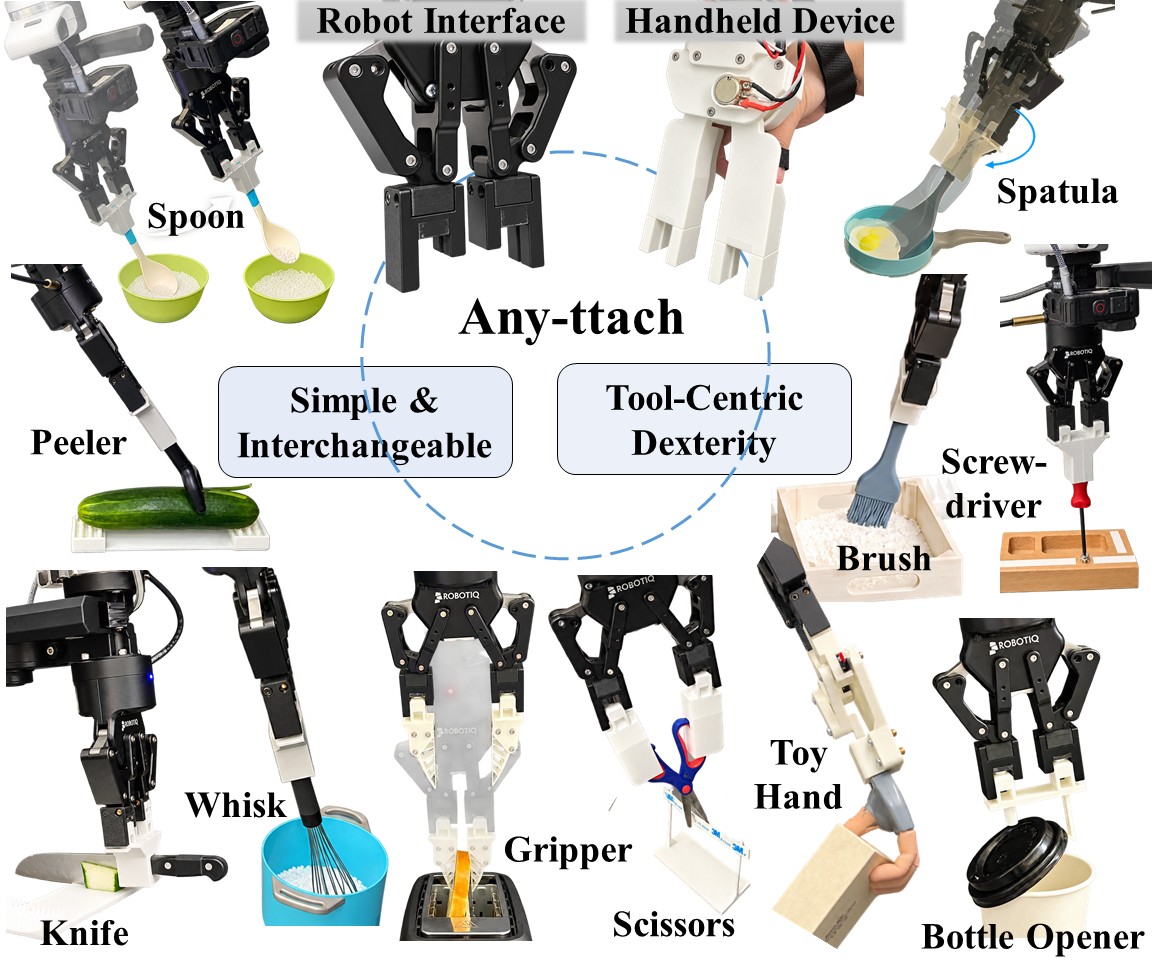}
    % \vspace{-0.5em}
    \caption{\textbf{Tool-centric design achieves dexterity and end-effector swapping with simplicity.}
    Any-ttach enables robots to perform diverse manipulation skills by rapidly switching between interchangeable tool modules through a standardized mechanical interface. By externalizing task-specific contact geometry into tools, the system reformulates manipulation dexterity as tool selection and skill execution rather than complex end-effector control.}
    \label{fig:hardware}
    % \vspace{-5pt}
\end{figure}

This gap motivates our alternative view: manipulation capability need not come only from increasing end-effector complexity. 
The functional role of human hands offers an evolutionary perspective: much of the dexterity comes from enabling tool using and tool making~\cite{marzke2013tool,marzke1997precision}. 
% Tools expand manipulation capability by embedding task-specific contact geometry into simple physical interfaces, rather than requiring all dexterity to reside in the hand itself. 
Tools expand manipulation capability by providing unique contact geometry specific to the tasks themselves, allowing a simple physical interface to perform interactions that would otherwise require more complex morphology and control.
This view aligns with extrinsic dexterity~\cite{dafle2014extrinsic}, where simple end-effectors achieve diverse behaviors by exploiting environmental constraints, object geometry, and contact interactions. 
Both perspectives point to a tool-centric route: robots can expand capability by choosing the right physical interface for each interaction and switching between tools and end-effectors as the task changes to achieve manipulation dexterity.

We propose \textbf{\textit{Any-ttach}}, a manipulation system built around an unactuated automatic mechanical interface that enables a 1-DoF actuator to directly attach and exchange everyday tools. Instead of relying on dexterous grasping or permanent attachment to use tools, \textit{Any-ttach} provides efficient automatic end-effector swapping and reliable mechanical attachment, supporting stable tool use in contact-rich interactions. The system has a hierarchical architecture in which a task planner decomposes high-level instructions into ordered tool–skill pairs, and skill modules execute each behavior in closed loop. These skills can be developed through learned policies, parameterized primitives, or motion planning. We also design a handheld tool interface that shares the same coupling mechanism to facilitate human demonstration collection and reduce the human–robot embodiment gap. 
% Through this design, we aim to explore a path of manipulation dexterity with simplicity: dexterity can emerge from composing skills across end-effectors and exploiting contact interactions within each skill, allowing even a simple 1-DoF actuator to achieve rich manipulation capabilities.

% Our key insight is that autonomous end-effector swapping can decouple manipulation dexterity from end-effector complexity, enabling a simple robotic interface to achieve diverse and long-horizon manipulation by composing existing tools and skills.
Through this system, we explored a path toward \textit{\textbf{manipulation dexterity with simplicity}}: autonomous end-effector swapping decouples dexterity from end-effector complexity, allowing even a simple 1-DoF interface to unlock diverse and long-horizon manipulation by composing task-specific tools and skills.
Our contributions are as follows:

\begin{itemize}

    % \item \textbf{A low-cost and extensible autonomous end-effector swapping system.}
    % A standardized interface with passive self-locking mechanism enables diverse end effectors to be integrated and autonomously exchanged while keeping tool-side hardware simple and actuator-free.
    \item \textbf{A low-cost and extensible hardware platform for autonomous end-effector swapping and tool use.}
    \textit{Any-ttach} combines passive automatic swapping, leverage actuation through an existing 1-DoF gripper, automatic tool-adapter generation, and handheld demonstration device. This design minimizes additional hardware and integration effort while facilitating the integration and extending of diverse end-effectors.

    % \item \textbf{A unified framework for tool-use manipulation.}
    % Tool use manipulation is decomposed into object-relative approach and local precise manipulation, supporting skills from robot trajectories, retargeted human demonstrations, and learned policies.
    \item \textbf{A unified and reusable framework for tool-use manipulation.}
    By separating object-tool alignment from local precise tool-use manipulation, the framework enables collected manipulation skills to be reused across changes in object pose while accommodating robot trajectories, retargeted human demonstrations, and learned policies within a shared execution structure.

    % \item \textbf{Extensive experimental validation and system-level demonstrations.}
    % Experiments demonstrate robust swapping, scalable end-effector integration, diverse tool-use skills, and autonomous composition of multiple tools and skills in long-horizon tasks.
    \item \textbf{Experimental evidence for expanding manipulation capability through tool and skill composition.}
    Experiments demonstrate reliable autonomous swapping and diverse tool-use skills acquired through different methods. Long-horizon multi-tool demonstrations further highlight how interchangeable tools and reusable skills expand manipulation capabilities without requiring a complex robotic hand.

\end{itemize}

%%%%%%%%%%%%%%%%%%%%%%%%%%%%%%%%%%%%%%%%%%%%%%%%%%%%%%%%%%%%%%%%%%%%%%%%%%%%%%%%

\section{Related Work}
\label{sec:related_work}

\subsection{Dexterity in Robotic Manipulation}

A dominant paradigm in robotic manipulation pursues intrinsic dexterity by embedding capability directly into the end-effector complexity.
High-DoF hands enable rich in-hand manipulation and functional grasping behaviors~\cite{andrychowicz2020learning,shaw2023leap,wang2022dexgraspnet,agarwal2023dexterous}, but typically introduce high-dimensional contact dynamics and substantial control complexity, 
and majorly focus on in-hand manipulation and grasping rather than more challenging tool-use tasks.
In contrast, \textit{extrinsic dexterity} shows that simple end-effectors can achieve rich behaviors by exploiting environmental structure, object geometry, and contact constraints~\cite{dafle2014extrinsic,zhou2022learninggraspungraspableemergent}. 
\textit{Any-ttach} follows a complementary direction.
Rather than increasing in-hand complexity or relying primarily on environmental fixtures, we externalize task-specific contact geometry into interchangeable tool modules.
This shifts part of manipulation dexterity from continuous control into physical reconfiguration, enabling a simple robot-side interface to access diverse interaction capabilities through end-effector swapping.

% Queue the comparison table for the top of the right column on page 2.
\begin{table}[!t]
\centering
\caption{Comparison of existing tool-changing systems}
\vspace{-1em}
\label{tab:tool_changer_comparison}
\scriptsize
\setlength{\tabcolsep}{2pt}
\renewcommand{\arraystretch}{1.0}
\setlength{\aboverulesep}{0.5pt}
\setlength{\belowrulesep}{0.5pt}
\renewcommand{\tabularxcolumn}[1]{m{#1}}
\begin{tabularx}{\columnwidth}{@{}>{\centering\arraybackslash}m{0.18\columnwidth}>{\raggedright\arraybackslash}m{0.27\columnwidth}>{\raggedright\arraybackslash}m{0.22\columnwidth}>{\centering\arraybackslash}m{0.16\columnwidth}>{\centering\arraybackslash}X@{}}
\toprule
\textbf{System}
& {\centering\textbf{Coupling mech.}\par}
& {\centering\textbf{Swap actuation}\par}
& \textbf{Daily (artic.)\newline tools}
& \textbf{Human\newline demo\newline device} \\
\midrule
ATI~\cite{ati_tool_changer}
& Cam-ball locking & Auto. (pneumatic) & -- (--) & -- \\
SCHUNK SWS~\cite{schunk_sws}
& Ball locking & Auto. (pneumatic) & -- (--) & -- \\
Dhakal~\cite{dhakal2019tool}
& Pin-slot bayonet & \textbf{Passive auto.} & -- (--) & -- \\
Cheong~\cite{cheong2024switchable}
& Switchable magnet & Auto. (dedicated motor) & -- (--) & -- \\
Song~\cite{song2024coaxial}
& Coaxial magnetic gear & \textbf{Passive auto.} & -- (--) & -- \\
Mourtzis~\cite{mourtzis2022design}
& Kinematic + threaded lock & \textbf{Passive auto.} & -- (--) & -- \\
Berenstein~\cite{berenstein2018open}
& Conical coupling + locking pins & \textbf{Passive auto.} & $\pmb{\checkmark}$ (--) & -- \\
ROMAN~\cite{li2022roman}
& Magnetic coupling & Auto. (dedicated motor) & $\pmb{\checkmark}$ ($\pmb{\checkmark}$) & -- \\
\midrule
\textbf{\textit{Any-ttach}} (Ours)
& Guided coupling + locking keys & \textbf{Passive auto.} & $\pmb{\checkmark}$ ($\pmb{\checkmark}$) & $\pmb{\checkmark}$ \\
\bottomrule
\end{tabularx}
\end{table}

\subsection{Tool-use Manipulation}

Robot tool use extends manipulation capability beyond the native end-effector geometry and has been studied through tool affordance reasoning, skill learning, compliant control, and tool selection~\cite{kemp2007challenges,ren2023leveraging,chen2025tool,orsula2025learning,dong2025robustness}.
Many existing systems use tools by grasping them with a general-purpose gripper or anthropomorphic hand.
This introduces a robot--tool contact interface in addition to the tool--object interaction, where external forces during manipulation can cause tool slip, rotation, and pose variability~\cite{bicchi2000hands,holladay2019force}.
Such variability can reduce execution repeatability and complicate demonstration collection and policy learning.
% Such variability can reduce execution repeatability and complicate demonstration collection or policy learning when the assumed tool configuration changes across trials.
% Existing approaches address this through more dexterous hands or task-specific mounted end effectors, trading mechanical simplicity, stability, and flexibility.
% \textit{Any-ttach} instead mechanically constrains the robot--tool interface while preserving autonomous tool exchange, providing a repeatable tool configuration across manipulation skills.

Alternatively, tools can be directly mounted as fixed end effectors, providing stable connections but limiting flexibility when tasks require different tools. Hierarchical planning and learned manipulation policies have enabled complex manipulation through task decomposition, motion planning, and skill execution~\cite{ahn2022can,huang2023voxposer,chi2025diffusion}, but are generally evaluated with fixed robot end-effector configurations.
In contrast, \textit{Any-ttach} combines mechanically constrained tool attachment with autonomous end-effector swapping, enabling both physical interfaces and manipulation skills to be selected and composed across long-horizon tasks. This provides a platform for investigating hierarchical planning and diverse skill-learning methods across interchangeable end effectors and different tool-use behaviors.

\subsection{Tool-Changing Systems}

Tool changers enable automatic end-effector exchange
through various coupling mechanisms.
Commercial systems typically employ pneumatic locking
mechanisms~\cite{ati_tool_changer,schunk_sws},
while other designs achieve passive swapping through robot
motion~\cite{dhakal2019tool,song2024coaxial,mourtzis2022design,berenstein2018open}
or use dedicated motors for magnetic coupling~\cite{cheong2024switchable,li2022roman}.
Beyond conventional robotic end effectors,
Berenstein et al.~\cite{berenstein2018open} developed a passive
tool changer accommodating everyday tools, while
ROMAN~\cite{li2022roman} introduced mechanisms for manipulating
everyday and articulated tools.
As summarized in Table~\ref{tab:tool_changer_comparison}, \textit{Any-ttach} integrates passive automatic swapping, extensible support for everyday and articulated tools,
and a shared handheld demonstration device, enabling existing learning from human demonstration methods to be extended to diverse tool-use skills.

%%%%%%%%%%%%%%%%%%%%%%%%%%%%%%%%%%%%%%%%%%%%%%%%%%%%%%%%%%%%%%%%%%%%%%%%%%%%%%%%

\section{Tool-Centric Hardware Platform}
\label{sec:hardware}

The \textit{Any-ttach} hardware platform combines a standardized end-effector interface, a passive coupling mechanism, an extensible end-effector library, and a matching handheld demonstration device.
The shared mounting geometry supports autonomous tool exchange and provides a common tool interface for human demonstration and robot execution.

% Keep the hardware overview after the section opening, in the next column.
\suppressfloats[t]
\begin{figure}[t]
    \centering
    \includegraphics[width=0.98\linewidth]{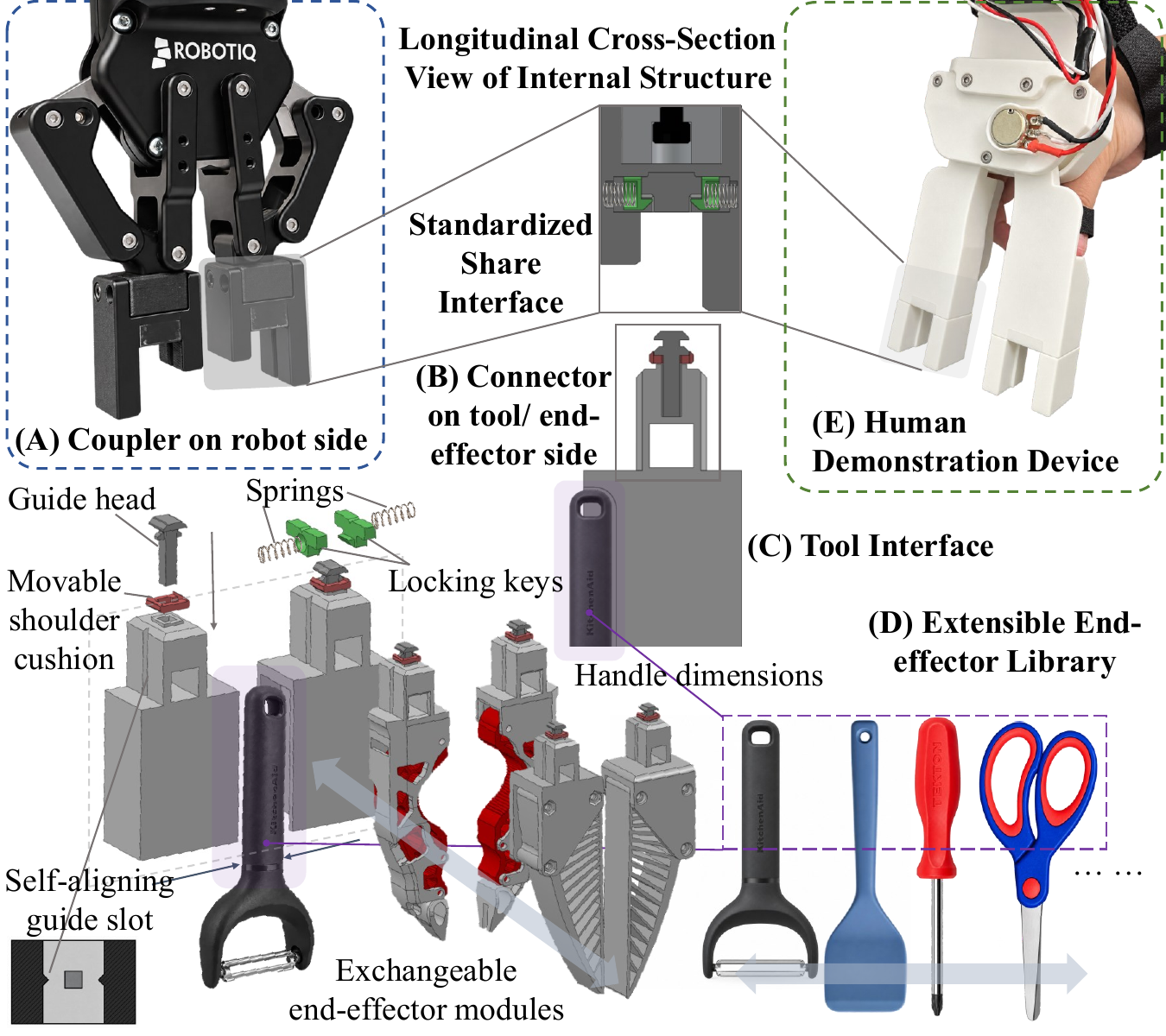}
    % \vspace{-1.0em}
    \caption{\textbf{Hardware Design.} 
    \textit{Any-ttach} uses a shared mechanical interface to couple diverse tools and end-effector modules to both the robot arm and the handheld demonstration device. 
    The system includes: 
    (A) a simple coupler on robot side with spring-loaded locking keys, 
    (B) an automatic end-effector swapping mechanism for passive locking and release, 
    (C) tool-side adapters that convert different handle dimensions into a standardized connector, 
    (D) an extensible end-effector library, and 
    (E) a corresponding finger-mounted demonstration device with the same interface. 
    } 
    \label{fig:hardware}
    % \vspace{-5pt}
\end{figure}

\subsection{Standardized End-Effector Interface}
\label{sec:interface}
The standardized interface provides a common mechanical connection between the robot and heterogeneous end-effector modules.
The design concentrates the locking mechanism on the robot side while keeping the tool-side adapter geometrically simple, allowing new tools to be integrated with minimal additional hardware.
The same interface supports passive attachment, mechanically triggered release, and mechanically constrained tool mounting through guided alignment and locking.

\textbf{Design Principles}
The interface design is guided by three principles:
(1) the tool-side connector should remain simple and inexpensive to fabricate, while reusable moving components are in the robot-side coupler;
(2) attachment and locking should be passive, eliminating the need for actuation on individual tool modules; and
(3) manipulation loads should not trigger unintended release, such that detachment requires a deliberate relative motion generated only when the tool is constrained by the dock.

\textbf{Mechanical Structure.}
As illustrated in Fig.~\ref{fig:hardware}, the interface consists of a robot-side coupler and a standardized tool-side connector.
The robot-side coupler contains two opposing spring-loaded locking keys constrained to move laterally.
The springs have a stiffness of $k=0.45~\mathrm{N/mm}$.
The keys are guided by complementary triangular features that constrain their motion to a single translational axis.
The coupler further incorporates a self-aligning guide slot that guides insertion and constrains transverse motion after engagement.
The tool-side connector includes a guide head with lead-in surfaces tapered at $45^\circ$ to the insertion axis and retention shoulders for mechanical engagement.
An axially movable shoulder cushion serves as the release element integrated into the tool-side connector and is used only during detachment.
Together, these components provide guided alignment, passive locking, and mechanically triggered release through a compact standardized interface.

\begin{figure}[t]
    \centering
    \begin{minipage}[b]{0.32\linewidth}
        \centering
        \parbox[c][1.82\linewidth][c]{\linewidth}{%
            \centering
            \makebox[\linewidth][c]{%
                \includegraphics[width=0.98\linewidth,height=1.815\linewidth,keepaspectratio]{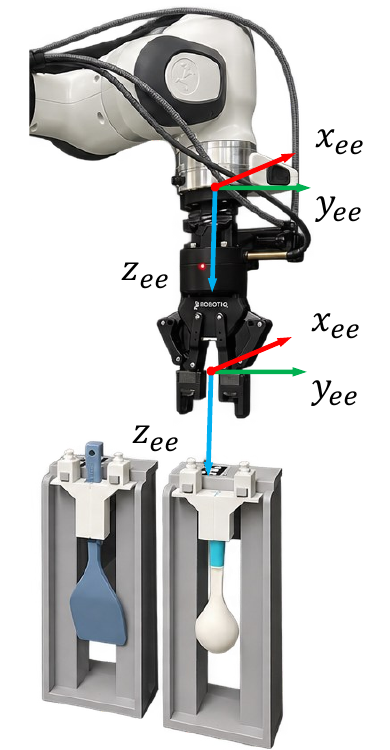}%
            }%
        }
        \par\smallskip
        {\footnotesize (a) Attachment}
    \end{minipage}\hspace{-0.02\linewidth}%
    \begin{minipage}[b]{0.32\linewidth}
        \centering
        \parbox[c][1.82\linewidth][c]{\linewidth}{%
            \centering
            \makebox[\linewidth][c]{%
                \includegraphics[width=1.10\linewidth,height=1.815\linewidth,keepaspectratio]{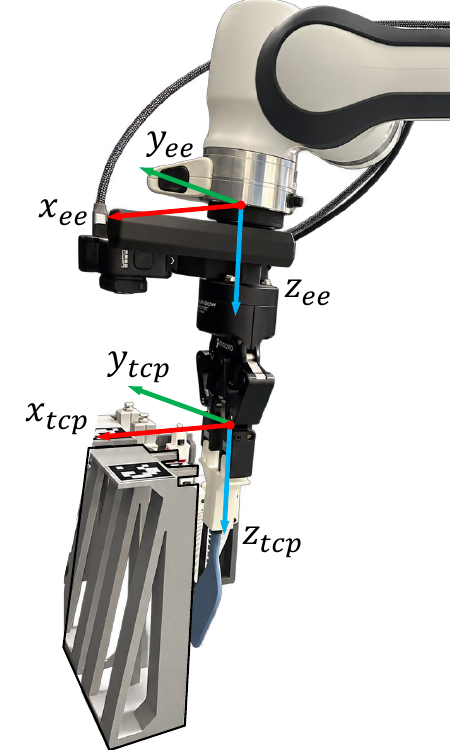}%
            }%
        }
        \par\smallskip
        {\footnotesize (b) Detachment}
    \end{minipage}\hspace{-0.02\linewidth}%
    \begin{minipage}[b]{0.32\linewidth}
        \centering
        \parbox[c][1.82\linewidth][c]{\linewidth}{%
            \centering
            \makebox[\linewidth][c]{%
                \includegraphics[width=1.10\linewidth,height=1.815\linewidth,keepaspectratio]{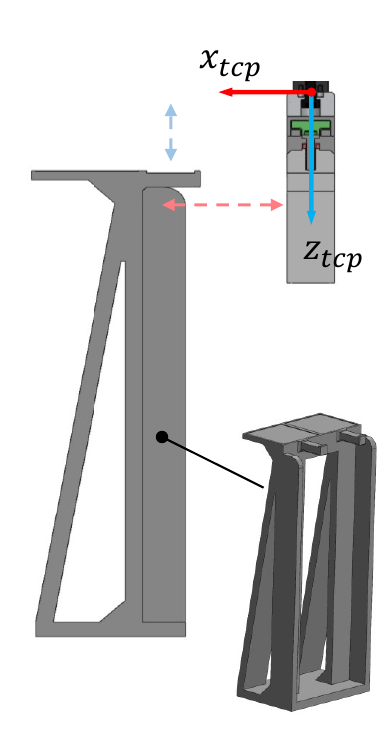}%
            }%
        }
        \par\smallskip
        {\footnotesize (c) Tool-shelf}
    \end{minipage}
    \par\medskip
    \makebox[\linewidth][c]{%
        \includegraphics[width=\linewidth]{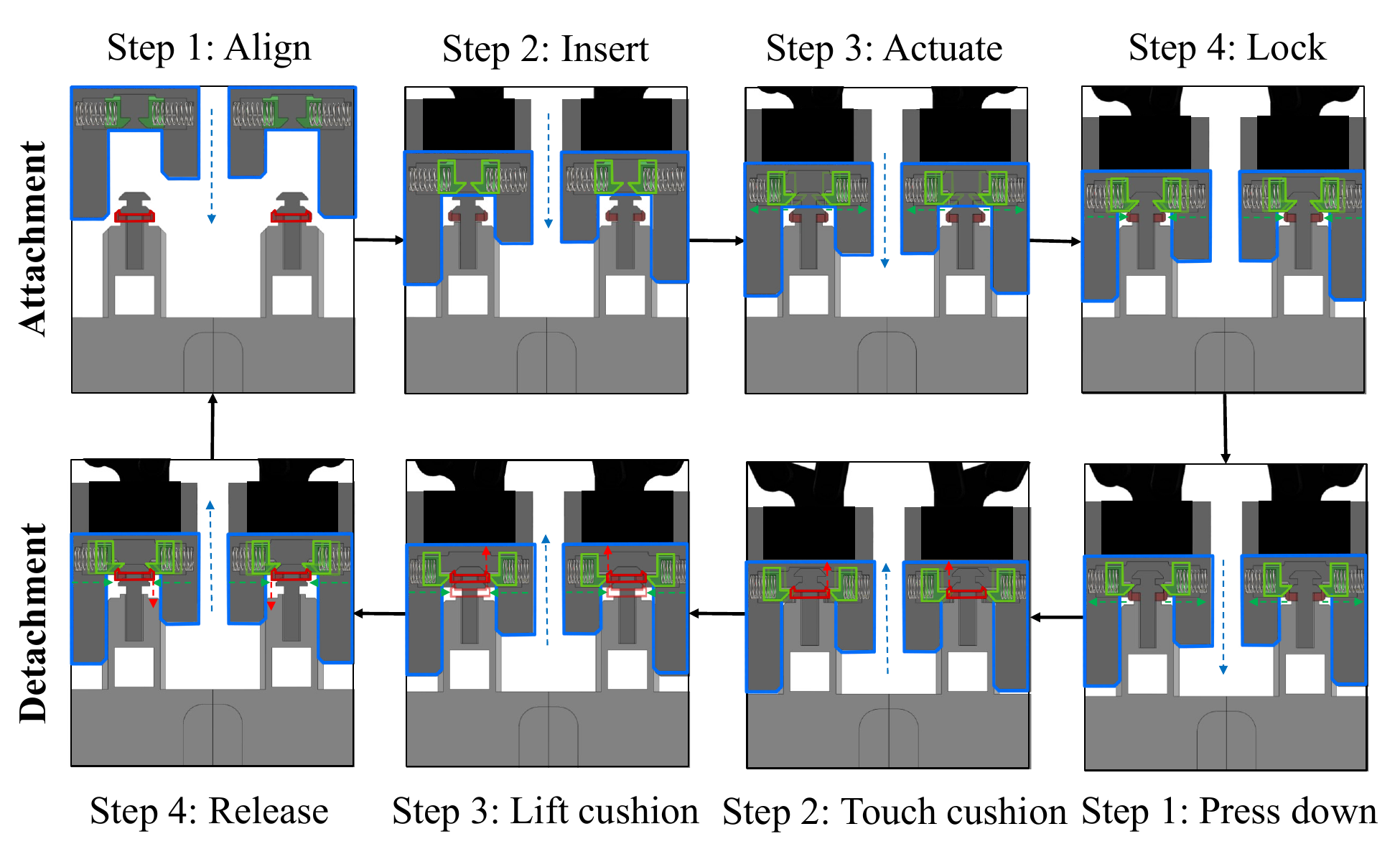}%
    }
    \par\smallskip
    \vspace{-0.8em}
    {\footnotesize (d) Attachment and detachment sequence}
    % \vspace{-0.5em}
    \caption{\textbf{End-effector Quick Swapping Configuration and Mechanism.}
    (a,b) Attachment and detachment configurations and reference frames.
    (c) Tool--shelf alignment.
    (d) Attachment (top) and detachment (bottom) sequences. The guide head passively cams the locking keys outward during insertion, after which they return to self-lock the tool. During detachment, shelf-constrained robot motion raises the movable shoulder cushion, retracting the keys and releasing the coupler.}
    \label{fig:mechanism}
    % \vspace{-5pt}
\end{figure}

\subsection{Quick Swapping Mechanism}
\label{sec:quick_swapping}
The quick swapping mechanism is illustrated in Fig.~\ref{fig:mechanism}.
During attachment, the robot aligns the coupler with the docked tool and advances downward along the coupling axis ($+z$).
Tool-side guide head displaces the locking keys outward due to spring compression during insertion; once the retention shoulders pass the keys, the springs drive the keys inward to lock the tool in place.
Robot compliance accommodates small residual pose errors during docking.
For detachment, the robot first translates the attached tool module along the $+x$ axis into the dock until the tool-side connector is mechanically constrained by the shelf.
Starting from the locked configuration, the robot then moves the coupler downward along the coupling axis ($+z$) until the locking keys contact the outer edge of the movable shoulder cushion.
The robot subsequently reverses direction and moves upward along the coupling axis ($-z$), lifting the shoulder cushion relative to the guide head until their tapered surfaces become continuous.
The resulting extended tapered profile cams the spring-loaded locking keys outward against the spring preload, disengaging them from the retention shoulders.
Once the locking keys are fully retracted and with sufficiently large acceleration, the robot-side coupler can be withdrawn while the tool remains constrained in the dock.
The same sequence is repeated to acquire the next end-effector. More detailed mechanical characterization are elaborated in Sec.~\ref{sec:mech_chara}.

\subsection{Extensible End-Effector Library}
\label{sec:library}

\textit{Any-ttach} is designed as an extensible end-effector platform rather than specific collection of tools.
% It supports four classes of end effectors: \textit{robotic fingertips}, \textit{everyday tools}, \textit{articulated tools}, and \textit{unconventional end effectors}.
The end-effector library spans robotic fingertips, everyday tools, and unconventional end effectors, accommodating diverse mechanical structures, including articulated tools such as scissors and tongs.
All modules share the same standardized swapping interface despite differences in morphology, actuation, and interaction mode.
By separating the common coupling from tool affordance geometry, new end effectors can be integrated without modifying the robot-side mechanism.

% \textbf{Tool Module Auto Generation}
\textbf{Adapter Generation with Tool Mesh Reconstruction.}
To integrate an arbitrary everyday tool, \textit{Any-ttach} automatically generates a compatible tool module from reconstructed tool geometry and merges it with the standardized connector interface.
For each tool, we reconstruct the handle geometry from 82--194 multi-view RGB images using COLMAP-based SfM~\cite{schonberger2016sfm}, 2D Gaussian Splatting~\cite{huang2024_2dgs}, and TSDF fusion~\cite{curless1996volumetric}.
The handle volume $\mathcal{H}$ is extracted from the fused geometry and expanded by a clearance $\epsilon=0.75~\mathrm{mm}$ as $\mathcal{H}_{\epsilon}=\mathcal{H}\oplus\mathcal{S}_{\epsilon}$.
A solid mounting block $\mathcal{B}$ is then formed by extending the bounding box of $\mathcal{H}_{\epsilon}$ by $\delta=5~\mathrm{mm}$ on each side along all three axes, and the tool-specific mounting cavity is obtained by Boolean subtraction as $\mathcal{M}=\mathcal{B}\setminus\mathcal{H}_{\epsilon}$.
Finally, the resulting cavity $\mathcal{M}$ is aligned to the desired tool pose and merged with the standardized adapter as $\mathcal{A}_{\mathrm{tool}}=\mathcal{A}_{0}\cup{}^{A}\mathbf{T}_{H}\mathcal{M}$, where ${}^{A}\mathbf{T}_{H}\in \mathrm{SE}(3)$ specifies the handle pose relative to the adapter.

\subsection{Human Demonstration Device}
\label{sec:device}

To collect human tool-use demonstrations, we design a handheld device sharing the same \textit{Any-ttach} interface as the robot (Fig.~\ref{fig:hardware}).
Inspired from related handheld systems such as UMI~\cite{chi2024umi} and YUBI~\cite{ohkawa2026yubi}, the device directly use the same interchangeable tool modules used during robot execution.
It reproduces the opening motion of the Robotiq 2F-85, with a potentiometer measuring gripper opening and a VIVE tracker~\cite{kulozik2024evaluating} recording 6-DoF pose.
The shared mounting configuration eliminates the need to estimate a separate human--tool grasp transform during retargeting.

\begin{figure*}[t]
\centering
\includegraphics[width=0.9\textwidth,keepaspectratio]{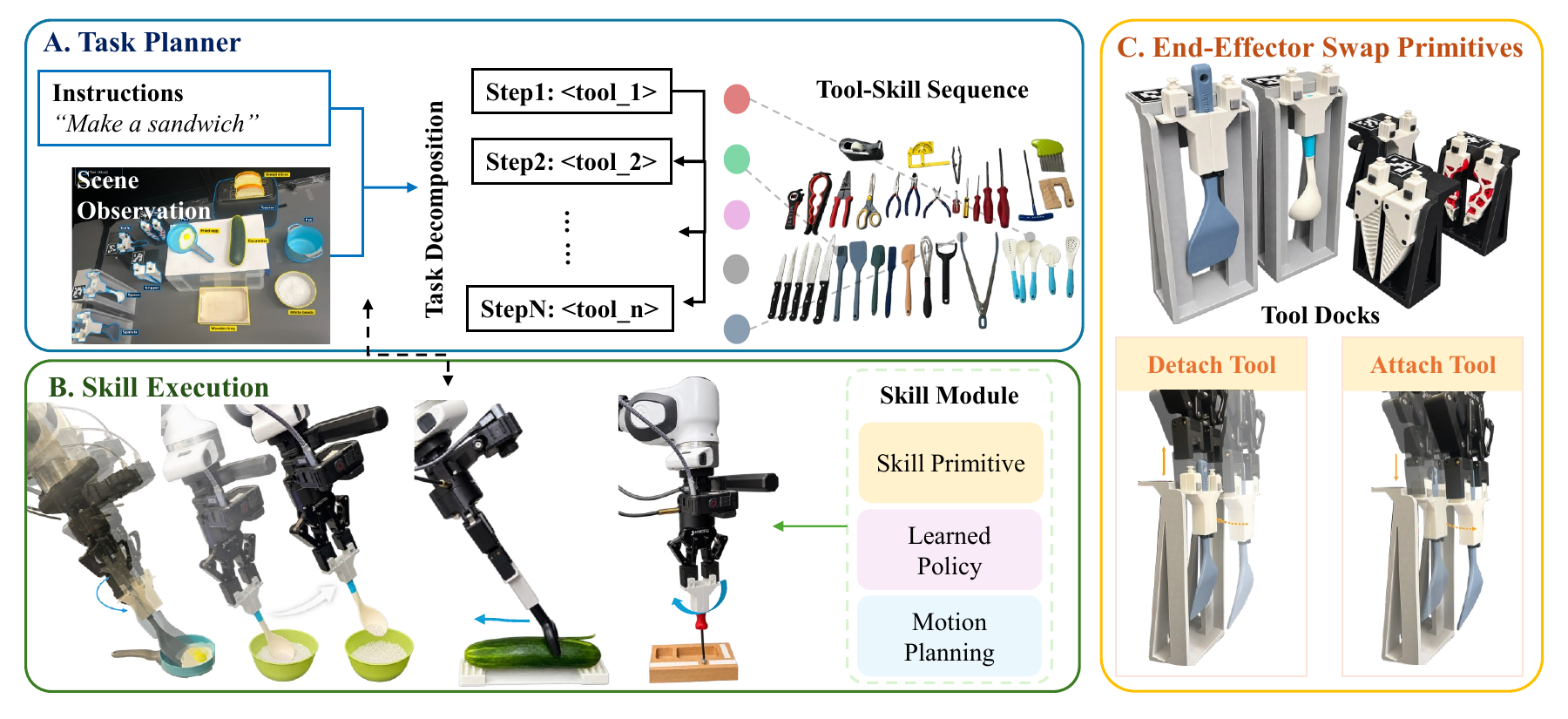}
\vspace{-1.0em}
\caption{\textbf{System pipeline of \textit{Any-ttach}.}
(A) A VLM decomposes the task into tool--skill pairs.
(B) Skill modules execute the selected manipulation behaviors.
(C) Swap primitives autonomously attach and detach the required end effectors.}
\label{fig:pipeline}
% \vspace{-5pt}
\end{figure*}

\section{Tool Manipulation Framework}
\label{sec:manipulation}

We represent each tool-use skill by an initial tool--object configuration and a subsequent manipulation behavior.
This allows skills acquired through different methods to share the same object-aligned execution framework.

\subsection{Tool-Centric Representation}
\label{sec:tool_representation}

For each task, we define an object frame $\mathcal{F}_{O}$ and a tool frame $\mathcal{F}_{T}$ from their observed geometry.
The initial relative tool--object configuration is defined as
${}^{\mathrm{Tool}}\mathbf{T}_{\mathrm{Object}}^{0}
=
({}^{\mathrm{Camera}}\mathbf{T}_{\mathrm{Tool}}^{0})^{-1}
{}^{\mathrm{Camera}}\mathbf{T}_{\mathrm{Object}}^{0}$.
Given a newly observed object pose, the desired tool pose is computed as
${}^{\mathrm{World}}\mathbf{T}_{\mathrm{Tool}}^{*}
=
{}^{\mathrm{World}}\mathbf{T}_{\mathrm{Object}}
({}^{\mathrm{Tool}}\mathbf{T}_{\mathrm{Object}}^{0})^{-1}$.
The corresponding end-effector pose is obtained through the calibrated tool-to-end-effector transform.
This object-relative representation allows the same skill to be initialized at different object poses.

\subsection{Tool Use Execution}
\label{sec:tool_execution}

Tool-use execution consists of two phases: \textit{Object Approach} and \textit{Precise Tool-Use Manipulation}.
During object approach, the robot moves the tool to the desired initial configuration computed from the current object pose.
During precise tool-use manipulation, the selected skill $\sigma$ generates robot motion and tool actuation commands as $(\mathbf{a}_t^{\mathrm{motion}},u_t^{\mathrm{tool}})=\sigma(t,\mathbf{o}_{\leq t};\mathbf{T}_{\mathrm{start}})$, where $\mathbf{T}_{\mathrm{start}}$ is the aligned TCP pose, $\mathbf{o}_{\leq t}$ denotes available observations, and $u_t^{\mathrm{tool}}$ represents optional tool actuation.
This separation allows skills acquired through different methods to share the same object-aligned execution framework.

\subsection{Skill Synthesis}
\label{sec:skill_synthesis}

We support three methods for obtaining tool-use skills: robot trajectory replay, human demonstration retargeting, and imitation learning.
All methods share the object-aligned execution framework but differ in how manipulation actions are generated.

\textbf{Robot Trajectory Replay.}
Given a robot demonstration
$\mathcal{D}_{R}=\{(\mathbf{P}_{R}(t),u_{R}(t))\}_{t=0}^{T}$,
where $\mathbf{P}_{R}(t)\in SE(3)$ denotes the recorded TCP pose and $u_{R}(t)$ denotes tool actuation when applicable, we extract the relative trajectory
$\Delta\mathbf{T}_{R}(t)
=\mathbf{P}_{R}(0)^{-1}\mathbf{P}_{R}(t)$.
During execution, the recorded motion is transferred to the object-aligned start configuration as
$\mathbf{P}_{R}^{*}(t)
=\mathbf{T}_{\mathrm{start}}\Delta\mathbf{T}_{R}(t)$,
while preserving the corresponding actuation commands.

\textbf{Human Demonstration Retargeting.}
Human demonstrations are collected using our handheld device, which shares the same tool mounting configuration as the robot.
Let ${}^{H}\mathbf{T}_{V}(t)$ denote the tracked VIVE pose in the demonstration frame and ${}^{V}\mathbf{T}_{\mathrm{TCP}}$ the fixed tracker-to-TCP calibration.
The demonstrated TCP trajectory is recovered as
$
\mathbf{P}_{H}(t)
=
{}^{H}\mathbf{T}_{V}(t)
{}^{V}\mathbf{T}_{\mathrm{TCP}}.
$
To reduce tracking noise while preserving the demonstrated motion, we apply a bounded smoothing operator
$\hat{\mathbf{P}}_{H}=\mathcal{S}_{\epsilon}(\mathbf{P}_{H})$,
subject to
$
\|\hat{\mathbf{p}}_{H}(t)-\mathbf{p}_{H}(t)\|_2
\leq 2~\mathrm{mm}, 
 d_{SO(3)}(\hat{\mathbf{R}}_{H}(t),\mathbf{R}_{H}(t))
\leq 0.5^\circ,
$
where $d_{SO(3)}$ measures rotational deviation.
The resulting trajectory is normalized and retargeted to the robot as
$
\mathbf{P}_{H}^{*}(t)
=
\mathbf{T}_{\mathrm{start}}
\hat{\mathbf{P}}_{H}(0)^{-1}
\hat{\mathbf{P}}_{H}(t).
$
Since demonstrations and execution share the same robotic embodiment and tool configuration, no cross-embodiment retargeting is required.

\textbf{Learning-Based Policy.}
For tasks requiring closed-loop manipulation, we collect
task-specific demonstrations through robot teleoperation,
recording wrist-view RGB images, 6-DoF tool-interface poses,
and gripper states alongside the corresponding actions.
We train an imitation learning policy, implemented using
Diffusion Policy (DP)~\cite{chi2025diffusion},
to predict robot motion and tool-actuation commands from
observations
$o_t=(I_t,{}^{W}\mathbf{T}_{\mathrm{interface},t},g_t)$.
During deployment, the policy predicts an action sequence
$\mathbf{A}_t=\pi_{\theta}(o_t)$ and executes it in a
receding-horizon manner using updated observations,
enabling learning-based policies within the framework.

%%%%%%%%%%%%%%%%%%%%%%%%%%%%%%%%%%%%%%%%%%%%%%%%%%%%%%%%%%%%%%%%%%%%%%%%%%%%%%%%
\subsection{Hierarchical Planning and Execution}
\label{sec:planning}

For long-horizon tasks, \textit{Any-ttach} coordinates tool selection, end-effector exchange, and manipulation skills through a hierarchical execution pipeline.
Given a task instruction, the system generates an ordered sequence of tool--skill pairs
$
\mathcal{P}=[(z^{(k)},\sigma^{(k)})]_{k=1}^{K},
$
where $z^{(k)}$ denotes the end effector selected for step $k$ and $\sigma^{(k)}$ denotes the corresponding manipulation skill.

\textbf{Task-Level Planning.}
A vision--language model~\cite{gpt5} receives the task instruction, current scene observation, and available tool and skill libraries, and decomposes the task into executable subtasks by selecting the required tool and skill for each step.
This converts a long-horizon instruction into a sequence of short-horizon tool-use behaviors.

\textbf{Runtime Execution and Verification.}
For each planned step, the system ensures that the required end effector is attached, performs an autonomous exchange when necessary, and executes the selected skill using the procedure described above.
A VLM-based monitor then evaluates the resulting scene against the intended subtask outcome.
If the post-condition is not satisfied, the skill is retried up to three times; otherwise, execution proceeds to the next tool--skill pair.
%%%%%%%%%%%%%%%%%%%%%%%%%%%%%%%%%%%%%%%%%%%%%%%%%%%%%%%%%%%%%%%%%%%%%%%%%%%%%%%%

\section{Experiments}
\label{sec:experiments}

We evaluate \textit{Any-ttach} through autonomous end-effector swapping (Sec.~\ref{sec:swapping_performance}), extensibility to diverse end effectors (Sec.~\ref{sec:extensibility}), and tool-use manipulation capabilities (Sec.~\ref{sec:capability}).
We further demonstrate the integration of these capabilities through two multi-tool long-horizon applications.

\subsection{Experimental Setup}

We evaluate \textit{Any-ttach} on a Franka Research 3 equipped with a Robotiq 2F-85, wrist-view camera, third-view RGB-D camera, and tool dock.
Experiments use 15 interchangeable end-effector modules spanning swapping, tool integration, manipulation, and long-horizon tasks. All modules share the same standardized interface.

\subsection{Autonomous End-Effector Swapping}
\label{sec:swapping_performance}
\subsubsection{Swapping Consistency and Disturbance Tolerance}

% We evaluate swapping using two experiments: repeated attachment--detachment cycles under fixed docking conditions (reliablity) and attachment--use--detachment trials after actual manipulation (robustness).
We evaluate autonomous swapping in two settings. First, we perform repeated attachment–detachment cycles under fixed docking conditions to assess swapping consistency over consecutive operations. Second, we perform attachment–disturbance–detachment trials across different end effectors, where external disturbances are applied to the attached tools to evaluate swapping performance after tool-side perturbations.

% \textbf{Repeated Swapping.}
% We perform 70 consecutive cycles with the GEAR Gripper, which produces the largest gravity-induced moment $\tau_g=mgd$ among the evaluated end effectors.
% The system achieves 100\% attachment and 98.57\% detachment success, with a single failure caused by accumulated docking pose drift during the repetition.
\textbf{Repeated Swapping.}
We perform 70 consecutive attachment--detachment cycles using the GEAR Gripper, which imposes the largest gravity-induced moment $\tau_g=mgd$ among the tested end-effectors.
Given the shared connector geometry, the guide contact force scales as $N\approx mgd/\ell$, where $\ell$ is the effective contact spacing.
This increases frictional loading and the risk of binding during swapping, making the GEAR Gripper the most demanding case in terms of gravitational loading.
The system achieves 100\% attachment and 98.57\% detachment success.

% \textbf{Swapping with Tool-Side Disturbances.}
% We further evaluate attachment--use--detachment cycles across all other end effectors, exposing the interface to residual pose variations introduced by actual manipulation.
% As shown in Table~\ref{tab:swapping_performance}, the average attachment and detachment success rates are 95.99\% and 96.35\% respectively, with all tested end-effectors achieve a complete rate above 91.67\%. These results show that the swapping mechanism remains reliable under pose variations introduced by tool use, supporting repeated attachment and reuse of the same end-effector modules across task cycles.
\textbf{Swapping Across Tools.}
We further evaluate attachment--disturbance--detachment cycles across the remaining end effectors, with external disturbances applied to the attached tools before detachment.
As shown in Table~\ref{tab:swapping_performance}, the average attachment and detachment success rates are 95.99\% and 96.35\%, respectively, with both rates reaching at least 91.67\% for every tested end effector.
The reported execution times include the complete robot swapping sequence of alignment, docking, and withdrawal, performed at conservative speeds for safety.
The swapping phase alone takes 11.32~s for attachment and 7.69~s for detachment on average, excluding approach and retreat motions.

\begin{table}[!t]
\centering
\caption{Autonomous end-effector swapping performance}
\vspace{-1em}
\label{tab:swapping_performance}
\footnotesize
\setlength{\tabcolsep}{2pt}
\renewcommand{\arraystretch}{1.0}
\setlength{\aboverulesep}{0pt}
\setlength{\belowrulesep}{0pt}
\begin{tabularx}{0.98\linewidth}{@{}l|*{2}{>{\centering\arraybackslash}X}|*{2}{>{\centering\arraybackslash}X}@{}}
\toprule
\multicolumn{1}{c|}{\multirow{2}{*}{\textbf{Test Cycle}}} &
\multicolumn{2}{c|}{\textbf{Success rate}} &
\multicolumn{2}{c}{\textbf{Execution Time}} \\
& Attach & Detach & Attach & Detach \\
\midrule
Repeated Swapping & 100\% & 98.57\% & 22.25\,s & 10.75\,s \\
\midrule
Swapping Across Tools & 95.99\% & 96.35\% & 52.91\,s & 64.76\,s \\
\bottomrule
\end{tabularx}
\vspace{-1em}
\par\vspace{\floatsep}
    \centering
    \caption{Mechanical characterization of the swapping interface}
    \label{tab:mechanical_characterization}
\footnotesize
\vspace{-1em}
\setlength{\tabcolsep}{3pt}
\renewcommand{\arraystretch}{1.0}
\setlength{\aboverulesep}{0.5pt}
\setlength{\belowrulesep}{0.5pt}
    \begin{tabular}{@{}lccc@{}}
        \toprule
        \textbf{Metric} &
        \textbf{$x$} &
        \textbf{$y$} &
        \textbf{$z$} \\
        \midrule
        Attachment force
        & $-8.63\,\mathrm{N}$
        & $-2.35\,\mathrm{N}$
        & $\mathbf{+3.96}\,\mathrm{N}$ \\
        Detachment force
        & $-5.83\,\mathrm{N}$
        & $+2.39\,\mathrm{N}$
        & $\mathbf{-47.49}\,\mathrm{N}$ \\
        Axial load
        & --
        & --
        & $\mathbf{75.4}\,\mathrm{N}$ \\
        Bending moment
        & $\mathbf{1.61}\,\mathrm{N\cdot m}$
        & -- $^{\dagger}$
        & -- \\
        \bottomrule
    \end{tabular}
    \par\vspace{2pt}
    {\scriptsize\raggedright
    $^{\dagger}$ Along the $y$-axis, opposing sidewalls
    prevent lateral disengagement, so no load limit measured.\par}
\end{table}

\subsubsection{Mechanical Characterization}
\label{sec:mech_chara}

We characterize the operational forces and payload capacity of the swapping interface to quantify its mechanical operating range and verify that it can cover the loading conditions associated with most end-effector and tool modules commonly used in robotic manipulation.

% \paragraph{Attachment and Detachment Forces.}
% We measure the external forces during attachment and detachment at the selected spring stiffness $k=0.45~\mathrm{N/mm}$.
% As summarized in Table~\ref{tab:mechanical_characterization}, the peak axial force is $+3.96~\mathrm{N}$ during attachment and $-47.49~\mathrm{N}$ during detachment.
% The substantially larger detachment force results from driving the locking keys outward against the spring preload during release, whereas attachment relies on passive camming and self-locking.
\textbf{Attachment and Detachment Forces.}
We measure the axial forces required for attachment and detachment.
As summarized in Table~\ref{tab:mechanical_characterization}, the peak forces are $+3.96~\mathrm{N}$ and $-47.49~\mathrm{N}$, respectively.
The larger detachment force results from pushing the locking keys outward against the spring preload during release, whereas attachment relies on passive camming and self-locking.

% \textbf{Payload.}
% We further evaluated the maximum payload supported by the fully locked interface along the tool axis ($z$) and in the lateral directions ($x/y$). The payload was gradually increased until disengagement, excessive displacement, or structural failure occurred. The measured payload capacities are summarized in Table~\ref{tab:mechanical_characterization}.

\textbf{Load-Bearing Capacity.}
We further evaluate the load-bearing capacity of the fully locked structure under axial and off-axis loading.
As summarized in Table~\ref{tab:mechanical_characterization},
the measured limits are 75.4 N in axial loading and
1.61 N$\cdot$m in off-axis bending.
Loads beyond these limits may lead to unintended
detachment or structural damage. 
Since the current prototype is 3D-printed in PLA,
the load-bearing capacity could be further improved by adopting materials of higher strength.

\subsection{Extensibility to Diverse End-effectors}
\label{sec:extensibility}

% \textit{Any-ttach} supports 15 end-effector modules spanning robotic end effectors, everyday tools, articulated tools, and unconventional end effectors.
% Despite differences in DoF, compliance, size, mass, and interaction mode, all modules are compatible with the same robot-side swapping mechanism.
% All 15 modules used throughout our experiments were successfully integrated through the standardized connector without modifying the robot-side mechanism, demonstrating the extensibility of the interface.

We demonstrate the extensibility of \textit{Any-ttach} by integrating 15 end-effector modules spanning robotic end effectors, everyday tools (including an articulated tool), and unconventional end effectors.
Despite differences in mechanical structure and interaction modes, all modules use the same standardized connector without modifying the robot-side swapping mechanism.

We demonstrate that our handle reconstruction provides sufficient geometry for tool adapter generation without requiring accurate digital twin models, facilitating the integration of new tools.
We evaluate this on three physical YCB tools~\cite{calli2017ycb}: a fork, screwdriver, and wrench, reconstructing their handles from our own multi-view RGB images rather than using the available YCB reference models.
As shown in Table~\ref{tab:ycb_integration_study}, our reconstructed handles qualitatively agree with the reference geometry, and all generated adapters successfully fit the physical tools.

\begin{table}[!t]
\centering
\caption{Reconstruction-based tool-module generation}
\vspace{-1em}
\label{tab:ycb_integration_study}
\footnotesize
\setlength{\tabcolsep}{1pt}
\renewcommand{\arraystretch}{1.0}
\setlength{\aboverulesep}{0.5pt}
\setlength{\belowrulesep}{0.5pt}
\renewcommand{\tabularxcolumn}[1]{m{#1}}

% Equal-height cells show each reconstruction-to-assembly example as one row.
\newcommand{\integrationcell}[1]{\parbox[c][9.8mm][c]{\linewidth}{\centering #1}}
\begin{tabularx}{0.90\columnwidth}{@{}
    >{\centering\arraybackslash}m{0.18\columnwidth}
    >{\centering\arraybackslash}m{0.21\columnwidth}
    >{\centering\arraybackslash}m{0.16\columnwidth}
    >{\centering\arraybackslash}X@{}}
\toprule
\makecell{\textbf{YCB scan}\\\textbf{reference}} & \makecell{\textbf{Reconstructed}\\\textbf{handle}} & \makecell{\textbf{Generated}\\\textbf{interface}} & \makecell{\textbf{Assembly}\\\textbf{time (s)}} \\
\midrule
\integrationcell{\includegraphics[width=0.85\linewidth,height=9.2mm,keepaspectratio]{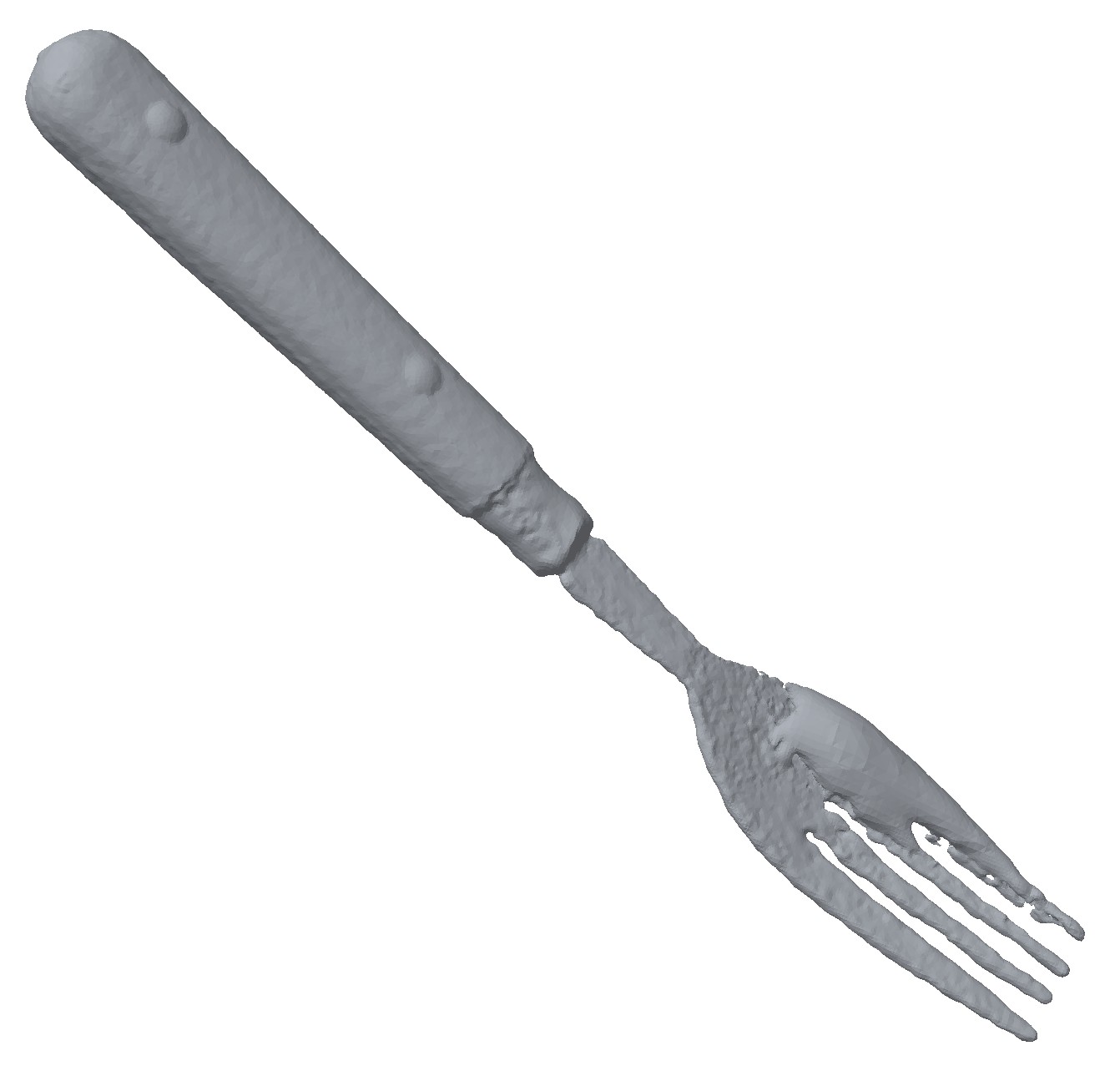}} &
\integrationcell{\includegraphics[width=0.85\linewidth,height=9.2mm,keepaspectratio]{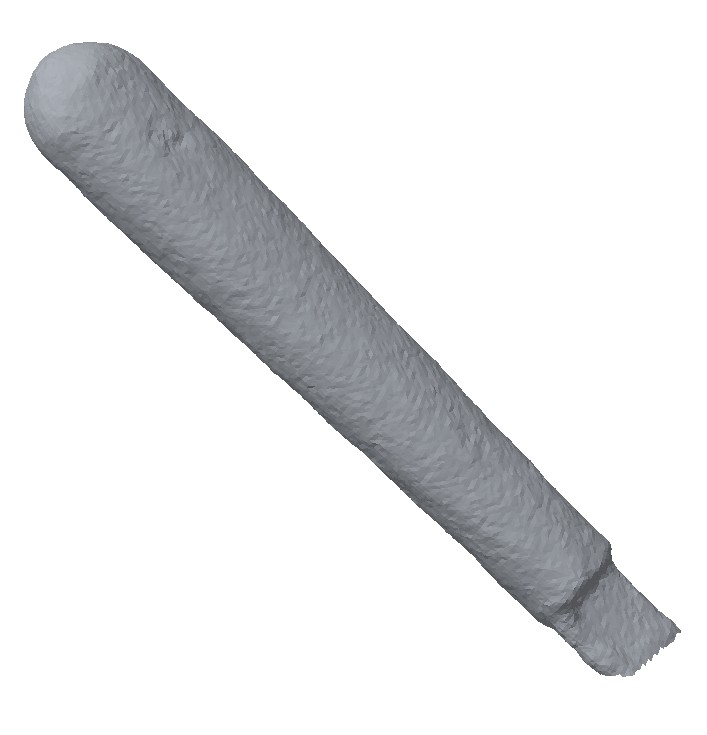}} &
\integrationcell{\includegraphics[width=0.85\linewidth,height=9.2mm,keepaspectratio]{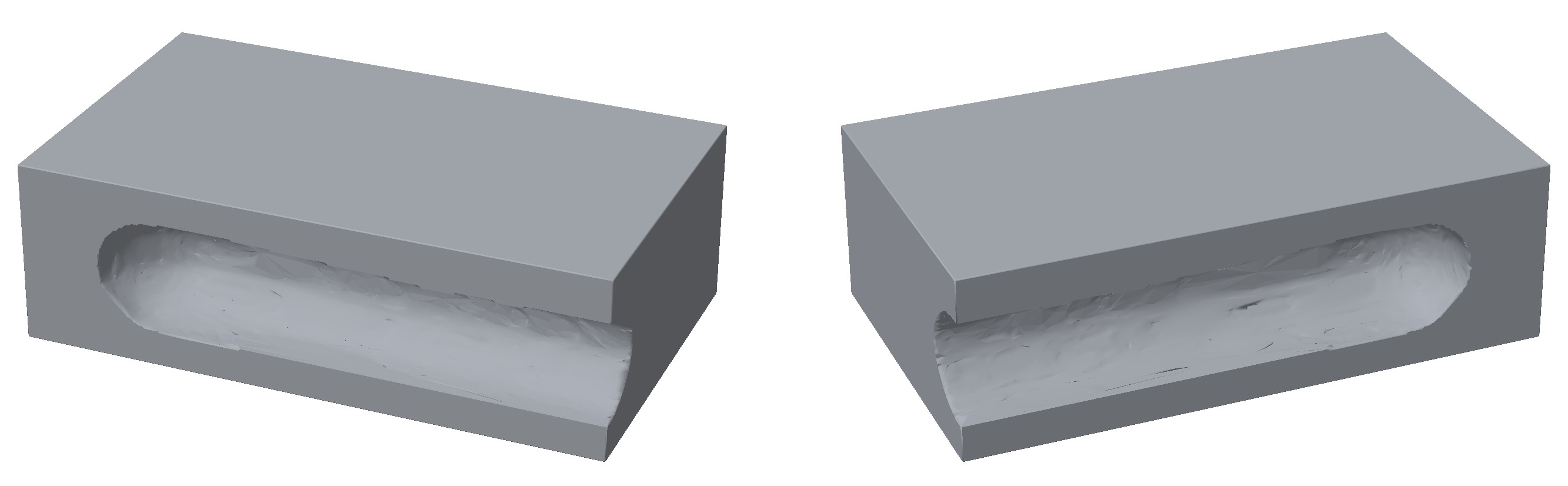}} &
\integrationcell{\makecell{\textit{Fork}\\$5.334 \pm 0.692$}} \\
\midrule
\integrationcell{\includegraphics[width=0.85\linewidth,height=9.2mm,keepaspectratio]{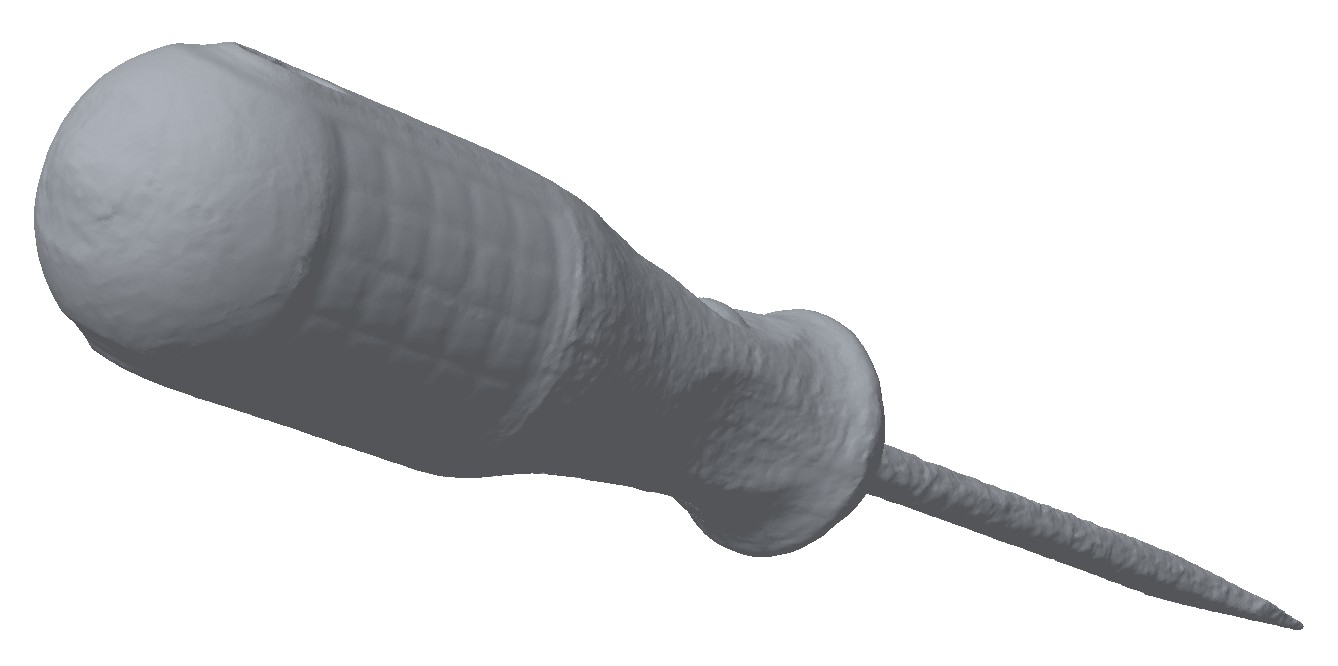}} &
\integrationcell{\includegraphics[width=0.85\linewidth,height=9.2mm,keepaspectratio]{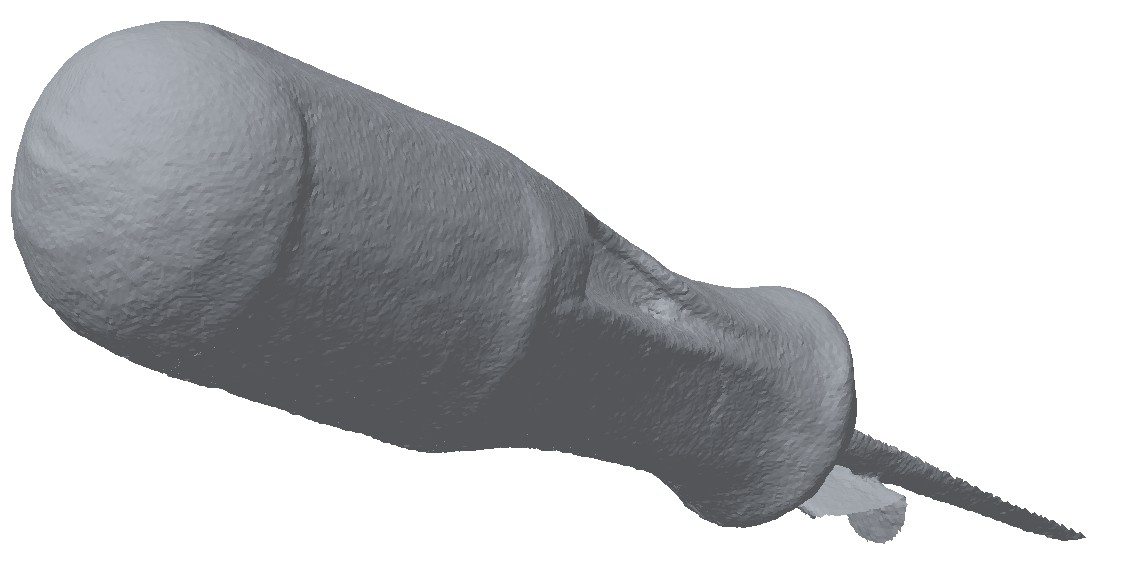}} &
\integrationcell{\includegraphics[width=0.85\linewidth,height=9.2mm,keepaspectratio]{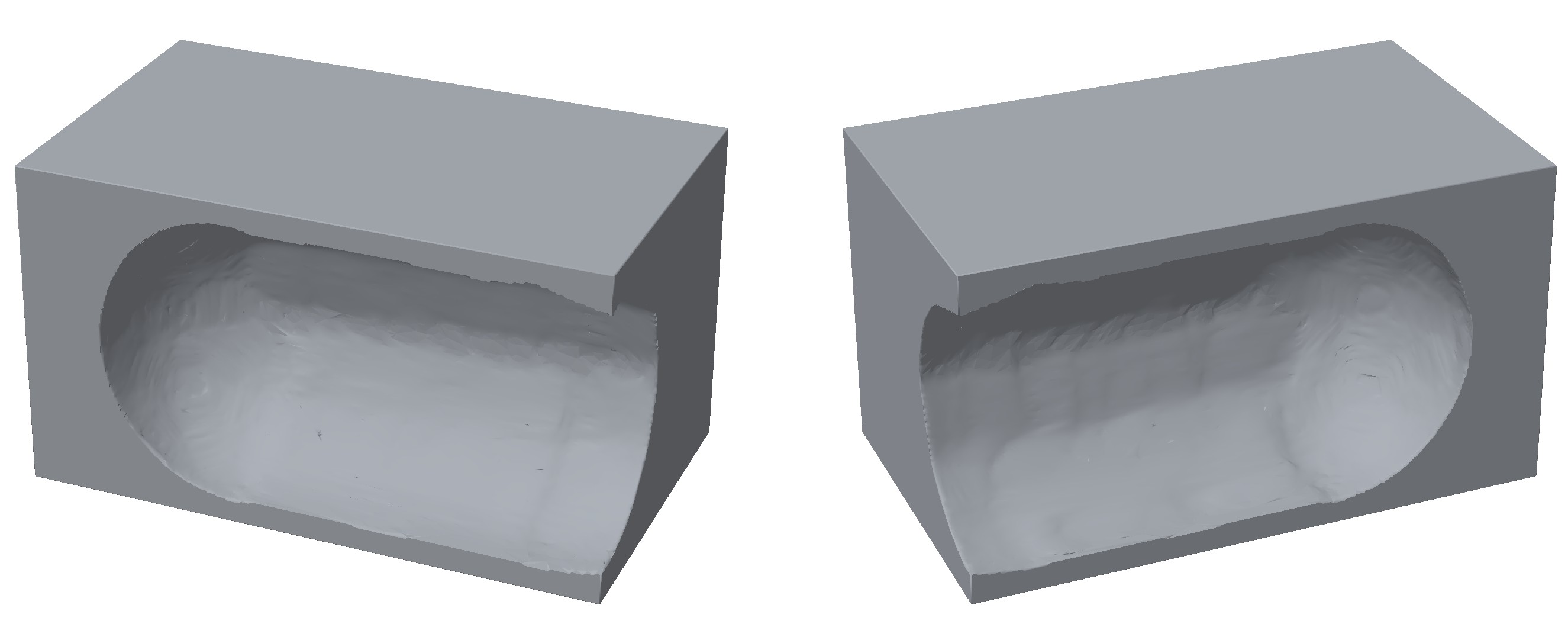}} &
\integrationcell{\makecell{\textit{Screwdriver}\\$6.726 \pm 0.713$}} \\
\midrule
\integrationcell{\includegraphics[width=0.85\linewidth,height=9.2mm,keepaspectratio]{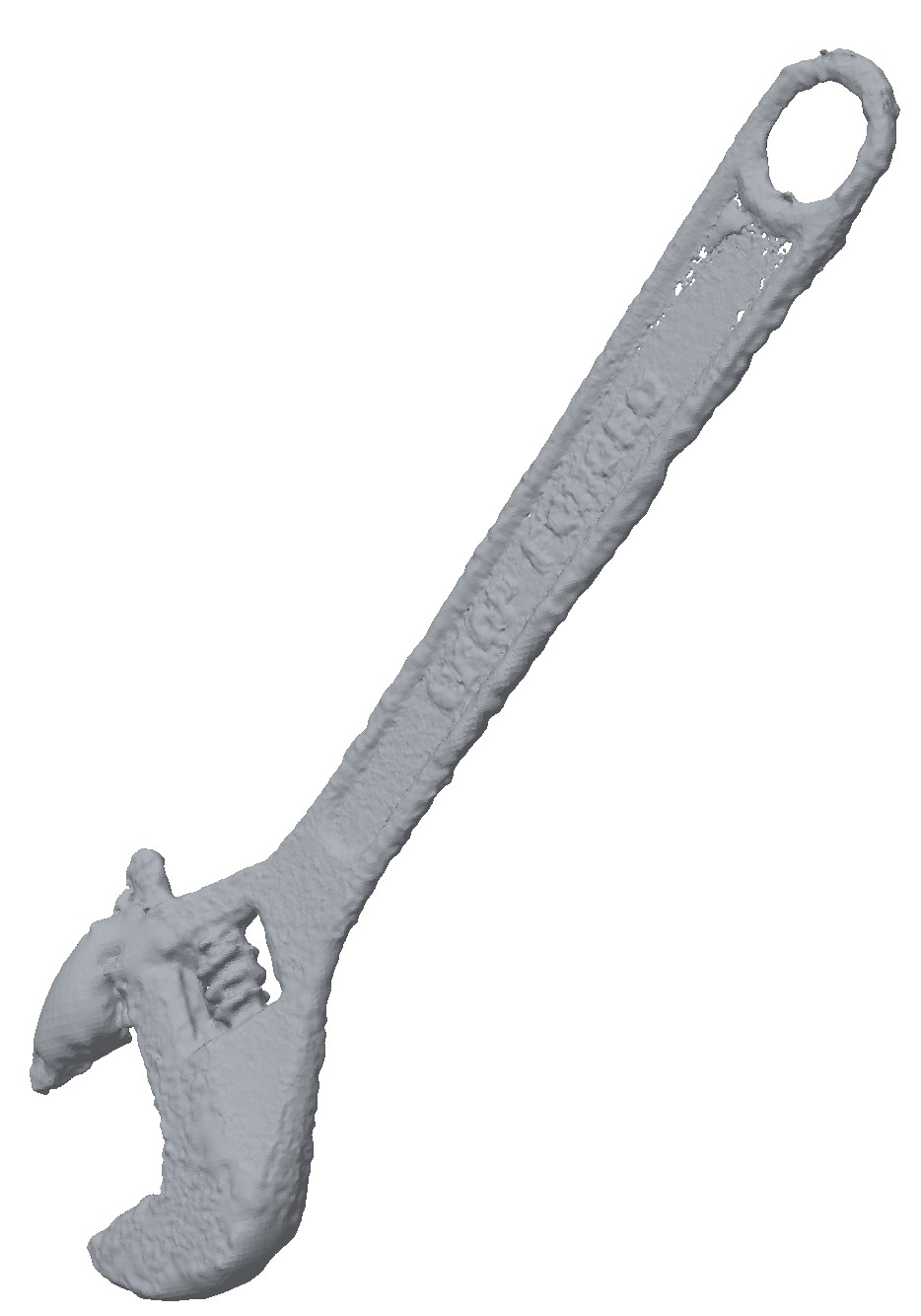}} &
\integrationcell{\includegraphics[width=0.85\linewidth,height=9.2mm,keepaspectratio]{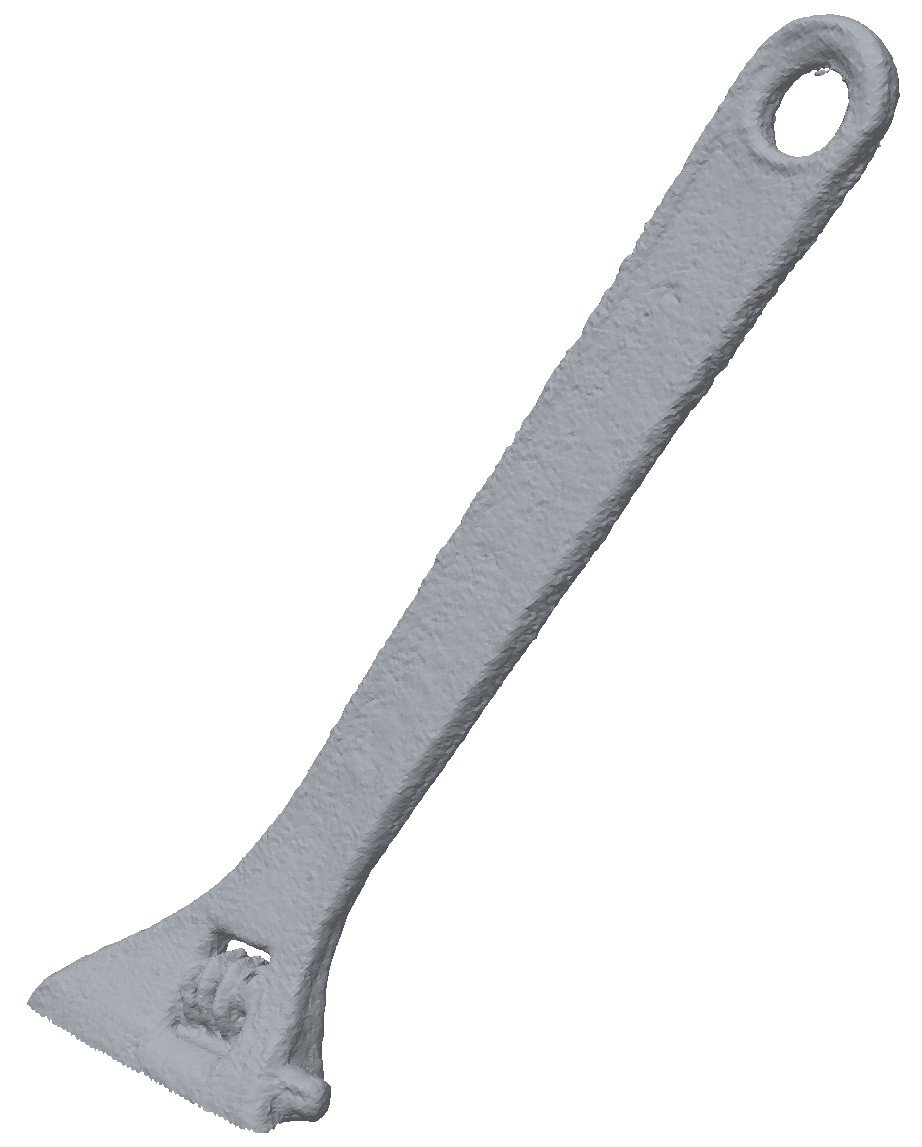}} &
\integrationcell{\includegraphics[width=0.85\linewidth,height=9.2mm,keepaspectratio]{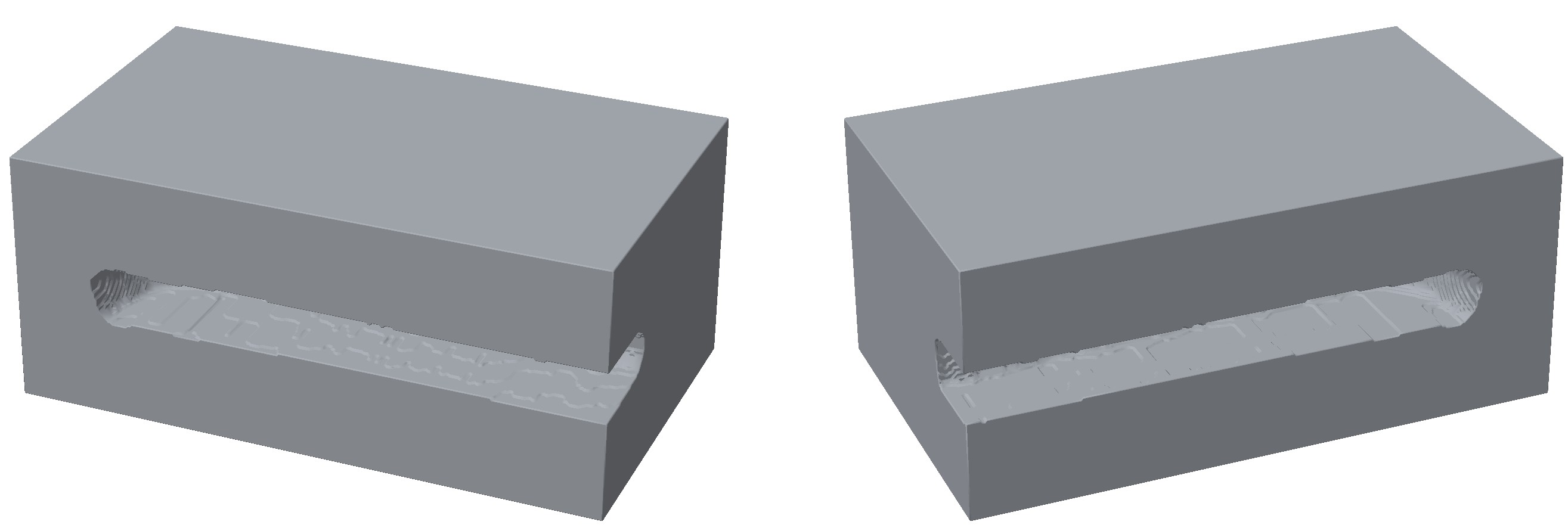}} &
\integrationcell{\makecell{\textit{Wrench}\\$12.404 \pm 1.376$}} \\
\bottomrule
\end{tabularx}
% \vspace{-5pt}
\end{table}

% Place Table IV before the manipulation text so it fits on page 6.
\begin{table}[!t]
\centering
\caption{Autonomous skill execution and tool interface comparison}
\vspace{-8pt}
\label{tab:skill_paradigm_performance}
\label{tab:interface_manipulation_performance}
\footnotesize
\setlength{\tabcolsep}{2pt}
\renewcommand{\arraystretch}{1.0}
\begingroup
\setlength{\tabcolsep}{1.5pt}
\renewcommand{\arraystretch}{1.0}
\setlength{\aboverulesep}{0pt}
\setlength{\belowrulesep}{0pt}
\begin{tabular*}{\columnwidth}{@{\extracolsep{\fill}}lcccccc@{}}
\toprule
\multicolumn{7}{c}{\textit{(a) Tool-use Skill Synthesis Methods}} \\
\midrule
\textbf{Method} & \textbf{Transfer} & \textbf{Scoop} & \textbf{Cut} & \textbf{Mix} & \textbf{Spread} & \textbf{Peel} \\
\midrule
Robot-data replay & 5/5 & 5/5 & 5/5 & 5/5 & 5/5 & 4/5 \\
\midrule
Human-data retarget & 2/5 & 3/5 & 4/5 & 5/5 & 5/5 & -- \\
\midrule
Diffusion Policy & 12/15 & 8/15 & 14/15 & 14/15 & 13/15 & 9/15 \\
\bottomrule
\end{tabular*}
\endgroup
\par\nointerlineskip
\begingroup
\setlength{\tabcolsep}{1.5pt}
\renewcommand{\arraystretch}{1.0}
\renewcommand{\tabularxcolumn}[1]{m{#1}}
\newcommand{\interfaceheader}[1]{\parbox[c][2\baselineskip][c]{\linewidth}{\centering\bfseries #1}}
% Remove rule gaps so the vertical separator remains continuous.
\setlength{\aboverulesep}{0pt}
\setlength{\belowrulesep}{0pt}
    \begin{tabularx}{\columnwidth}{@{}
    >{\centering\arraybackslash}m{0.19\columnwidth}
    >{\centering\arraybackslash}m{0.17\columnwidth}|
    >{\centering\arraybackslash}m{0.20\columnwidth}
    >{\centering\arraybackslash}m{0.12\columnwidth}
    >{\raggedright\arraybackslash}X@{}}
\multicolumn{5}{c}{\textit{(b) Tool Interface Comparison}} \\
\midrule
        \interfaceheader{Tool--Task} &
        \interfaceheader{Interface} &
        \interfaceheader{Task SR} &
        \interfaceheader{Manip.\\time} &
        \interfaceheader{Dominant failure/\\unstability} \\
        \midrule
        \multirow{2}{*}{\makecell{Spatula--\\Transfer}}
        & Grasp-based & 25\% (2/8) & 79.55\,s & {\scriptsize Handle slip} \\
        & \textit{Any-ttach} & \textbf{100\% (5/5)} & 75.32\,s & -- \\
        \midrule
        \multirow{2}{*}{Spoon--Scoop}
        & Grasp-based & 100\% (5/5) & 55.31\,s & {\scriptsize Shake in transfer} \\
        & \textit{Any-ttach} & 100\% (5/5) & 47.28\,s & -- \\
        \midrule
        \multirow{2}{*}{Brush--Spread}
        & Grasp-based & 100\% (5/5) & 40.00\,s & -- \\
        & \textit{Any-ttach} & 100\% (5/5) & 38.12\,s & -- \\
        \midrule
        \multirow{2}{*}{Peeler--Peel}
        & Grasp-based & \textbf{80\% (4/5)} & \textbf{44.33\,s} & {\scriptsize Handle stimulating slip} \\
        & \textit{Any-ttach} & 66.67\% (4/6) & 38.49\,s & {\scriptsize\mbox{Insufficient compliance}} \\
        \midrule
        \multirow{2}{*}{Knife--Slice}
        & Grasp-based & 0\% (0/5) & \multicolumn{2}{l@{}}{\scriptsize Slip under cutting torque} \\
        & \textit{Any-ttach} & \textbf{83.33\% (5/6)} & 86.61\,s & {\scriptsize Incomplete cut} \\
        \midrule
        \multirow{2}{*}{Cutter--Cut}
        & Grasp-based & 100\% (5/5) & 29.33\,s & -- \\
        & \textit{Any-ttach} & 100\% (5/5) & 27.21\,s & -- \\
        \midrule
        \multirow{2}{*}{Scissors}
        & Grasp-based & 0\% (0/5) & \multicolumn{2}{l@{}}{\scriptsize Grasp--actuation incompatibility} \\
        & \textit{Any-ttach} & 100\% (5/5) & 34.04\,s & -- \\
        \midrule
\multirow{2}{*}{\textbf{Average}}
        & Grasp-based & 57.86\% & 49.70\,s & \\
        & \textit{Any-ttach} & \textbf{92.86\%} & 45.28\,s &  \\
        \bottomrule
    \end{tabularx}
\endgroup
% \vspace{-5pt}
\end{table}

% Queue the long-horizon figure for the top of page 7.
\begin{figure*}[!t]
    \centering
    % Overlay panel labels without increasing the figure height.
    \newcommand{\taskpanel}[2]{%
        \begingroup
        \setbox0=\hbox{#1}%
        \leavevmode\rlap{\copy0}%
        \makebox[\wd0][l]{%
            \raisebox{\dimexpr\ht0-11pt\relax}[0pt][0pt]{%
                \hspace{2pt}\setlength{\fboxsep}{1.5pt}%
                {\transparent{0.55}%
                    \colorbox{white}{\transparent{1}\footnotesize\bfseries (#2)}%
                }%
            }%
        }%
        \endgroup
    }
    \begin{minipage}[c]{0.64\linewidth}
        \centering
        \setlength{\fboxsep}{0pt}
        \setlength{\fboxrule}{0.5pt}
        \fcolorbox{black}{white}{%
            \taskpanel{\includegraphics[width=\dimexpr\linewidth-2\fboxrule\relax,trim=0 3bp 0 4bp,clip]{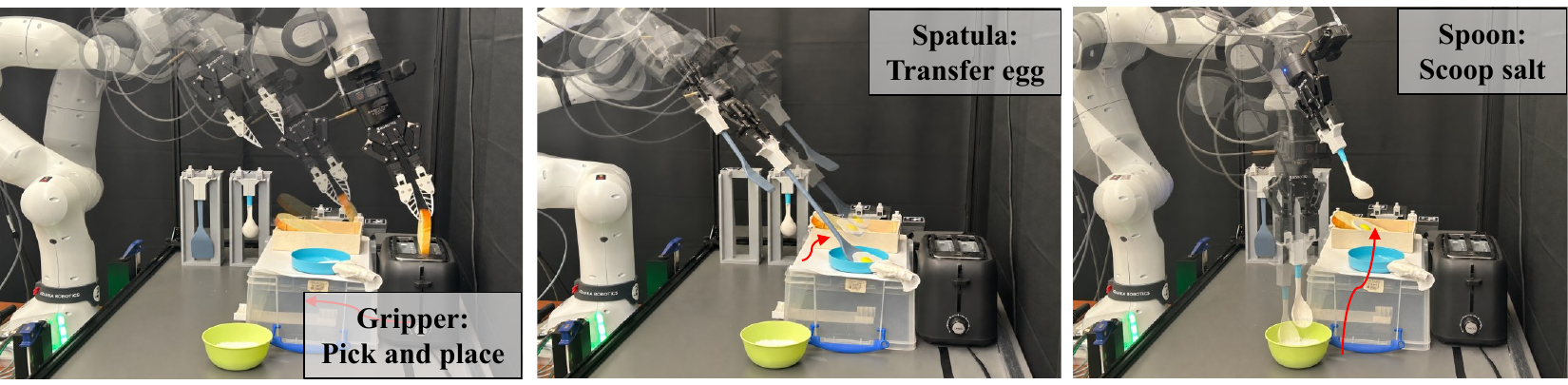}}{a}%
        }\par
        \vspace{2pt}
        \fcolorbox{black}{white}{%
            \taskpanel{\includegraphics[width=\dimexpr\linewidth-2\fboxrule\relax]{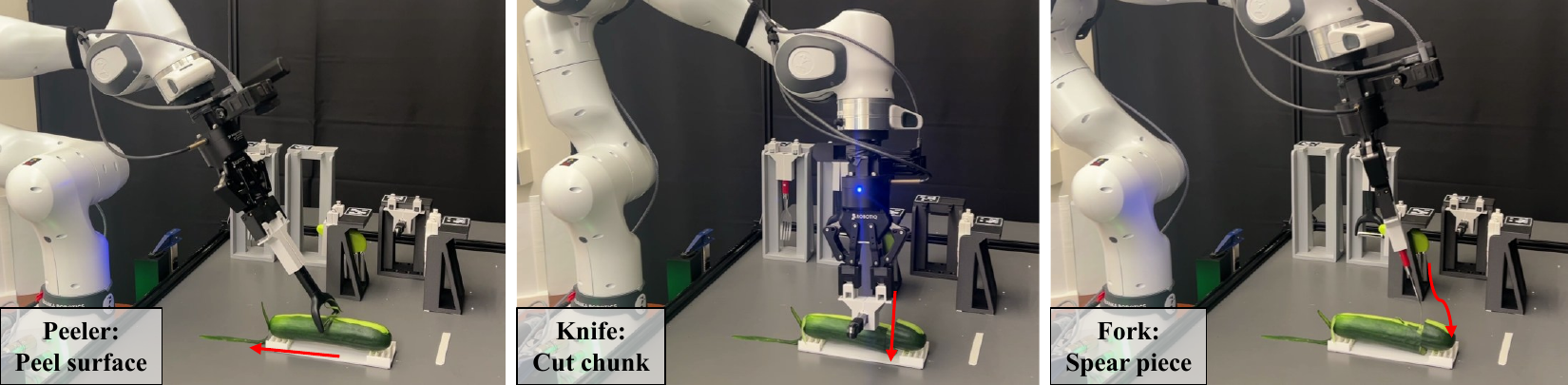}}{b}%
        }
    \end{minipage}\hfill
    \begin{minipage}[c]{0.34\linewidth}
        \centering
        \taskpanel{\includegraphics[width=\linewidth]{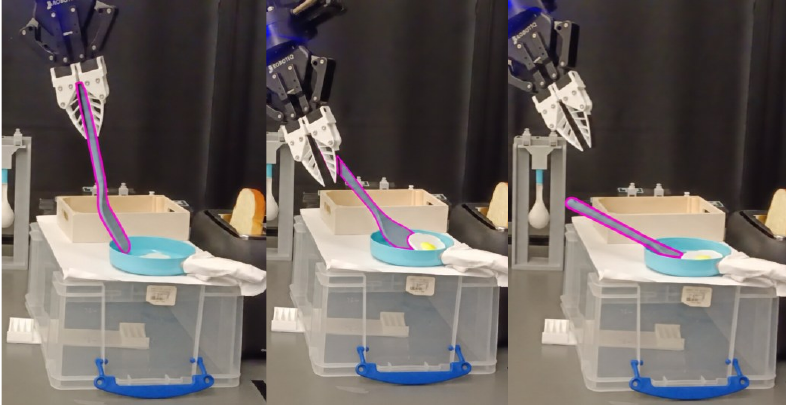}}{c}\par
        \vspace{2pt}
        \taskpanel{\includegraphics[width=0.45\linewidth,height=0.40\linewidth,keepaspectratio]{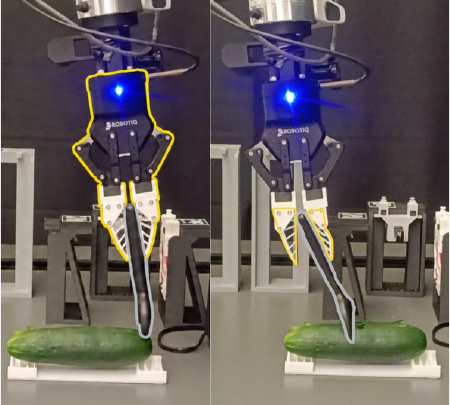}}{d}\hfill
        \taskpanel{\includegraphics[width=0.52\linewidth,height=0.40\linewidth,keepaspectratio]{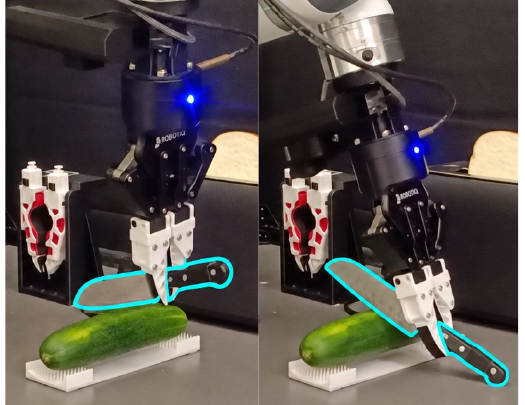}}{e}
    \end{minipage}
    \caption{\textbf{Qualitative Evaluation of Tool-Use Manipulation.}
    (a) \textbf{Sandwich Assembly:} The robot picks and places a bread with a gripper, transfers an egg with a spatula, and scoops salt with a spoon.
    (b) \textbf{Cucumber Preparation:} The robot peels the cucumber, cuts it into chunks, and spears a piece using a peeler, knife, and fork, respectively.
    (c--e) \textbf{Grasp-Based Tool Using:} Representative examples of (c) tool slippage during transferring, (d) relative tool motion during peeling, and (e) tool rotation during cutting, illustrating how contact forces can alter the tool configuration within a parallel gripper.
    }
    \label{fig:qualitative_evaluation}
    \label{fig:long_horizon}
    % \vspace{-5pt}
    
\end{figure*}

\subsection{Manipulation Capabilities}
\label{sec:capability}

\subsubsection{Tool-use Skills}
\label{sec:skill_paradigms}
Table~\ref{tab:skill_paradigm_performance}(a) compares autonomous execution of tool-use skills acquired through different skill synthesis methods under the shared object-aligned execution framework introduced in Sec.~\ref{sec:tool_representation}. After skill acquisition, the robot executes each skill autonomously during evaluation.
The comparison includes three methods: \textit{Robot-data replay}, which replays trajectories recorded on the robot; \textit{Human-data retarget}, which retargets human demonstrations collected with our handheld device; and \textit{diffusion policy} (DP)~\cite{chi2025diffusion}, which generates manipulation actions using a learned policy.

% \textbf{Skill implementation}
\textbf{Implementation Details.}
Each tool-use step is executed by a corresponding skill module.
For human-data retargeting, demonstrations collected with the handheld device are transformed to the robot TCP frame using the fixed device-to-TCP calibration, while robot-data replay uses trajectories recorded on the robot; both are executed relative to the same object-aligned initial configuration.
For learning-based skills, the observation consists of the workspace RGB image, the 6-DoF tool-interface pose in the robot frame, and the binary gripper state, while the policy outputs a receding-horizon sequence of end-effector actions.
Each diffusion policy is trained separately for 3,000 epochs on 30 demonstrations using a single NVIDIA GeForce RTX 5090 GPU.
% For pose-driven skills, the input is a target end-effector pose or waypoint sequence, and the output is a joint-space trajectory or scripted motion primitive.
% Tool coupling and decoupling follow predefined approach, docking, insertion/release, and retreat waypoints relative to the tool dock.
For skills based on trajectory, target end-effector poses or waypoints are converted into executable joint-space trajectories.
Tool attachment and detachment use manually programmed motion sequences consisting of approach, alignment, insertion/release, and retreat relative to the tool dock.

\textbf{Performance Analysis.}
We evaluate six representative tool-use tasks spanning distinct execution challenges, including translational and rotational motion, sustained contact, force transmission, and local manipulation. As shown in Table~\ref{tab:skill_paradigm_performance}(a), robot-data replay achieves the most consistent performance because the object-approach phase establishes the initial configuration and the recorded robot trajectory provides accurate local motion for precise manipulation. Human-data retargeting degrades on \textit{Transfer} and \textit{Scoop}, reflecting sensitivity to tracking, calibration, and cross-embodiment retargeting errors. DP performs well on visually and kinematically driven tasks but incomplete on \textit{Cut}; although it reaches the target region with 14/15 success, the absence of force/contact observations makes reliable forceful cutting difficult.

% \subsubsection{Manipulation Performance Across Interfaces}
% We characterize tool-use manipulation across two interface strategies: a parallel gripper, and our \textit{Any-ttach} interface. The comparison focuses on tool-use tasks spanning different interaction requirements, with the goal of revealing the capability profile and tradeoffs of each interface.

% We select representative tasks to span distinct interface-level interaction requirements. The benchmark includes low-load interaction, sustained contact, pose-sensitive manipulation, force/torque-intensive operation, articulated-tool use, and tasks requiring configuration flexibility. Together, these tasks probe different demands on tool retention, pose stability, wrench transmission, articulation, and relative tool configuration.

% Table~\ref{tab:interface_manipulation_performance}(b) evaluates the effect of the robot--tool interface on manipulation performance under teleoperation. For both interfaces, the same operator uses our handheld device to teleoperate the robot throughout each task. The grasp-based interface directly grasps the tool with the parallel gripper, whereas \textit{Any-ttach} mounts the same tool through the standardized connector. The robot, task objects, initial conditions, operator, and success criteria are kept consistent across the two interfaces.

\subsubsection{Manipulation Performance Across Interfaces}

We compare two robot--tool interfaces: a parallel gripper and our \textit{Any-ttach} interface.
When using the parallel gripper, the robot must maintain a stable gripper--tool contact while performing tool--object interactions.
Off-axis interaction forces and torques can cause tool slip or rotation within the grasp, changing the tool pose relative to the robot and complicating task execution, demonstration collection, and policy learning.
In contrast, \textit{Any-ttach} secures the tool through a mechanically constrained interface, avoiding the need to maintain a frictional grasp during tool--object interaction.

To evaluate these differences, both interfaces are tested under the same teleoperation setup, with identical task conditions and success criteria.
For the parallel gripper, we apply the maximum gripping force and manually select grasp configurations to ensure stable tool holding and proper orientation for each task.
As shown in Table~\ref{tab:interface_manipulation_performance}(b), \textit{Any-ttach} achieves an average success rate of 92.86\%, compared with 57.86\% for the parallel gripper across seven tasks.
Despite the manually configured grasps, the parallel gripper experiences tool slip during \textit{Transfer} (25\% vs. 100\%) and \textit{Slice} (0\% vs. 83.33\%) due to insufficient grasp retention under interaction loads.
Furthermore, the parallel gripper cannot simultaneously hold and actuate the tested scissors, whereas \textit{Any-ttach} achieves 100\% success.

For \textit{Peeling}, the parallel gripper achieves 80\% success compared with 66.67\% for \textit{Any-ttach}, as its Fin-Ray fingers provide passive compliance to accommodate curved-surface contact.
However, this structural compliance cannot prevent tool slip when interaction loads exceed grasp retention.
For \textit{Any-ttach}, controller-level compliance could improve adaptation to curved surface contact while preserving the stability of its mechanically constrained attachment. More qualitative observations are illustrated in Fig.~\ref{fig:qualitative_evaluation}(c--e).

\subsection{System-Level Application Demonstration}
\label{sec:application}

% We demonstrate the system-level capability of \textit{Any-ttach} through two multi-tool long-horizon manipulation tasks that require repeated end-effector exchange and composition of heterogeneous tool-use skills (Fig.~\ref{fig:long_horizon}). The two tasks comprise seven tool--skill steps in total: three in \textit{Sandwich Assembly} and four in \textit{Cucumber Preparation}.

% Examples to shrink the wording: 
% We demonstrate the system-level capability of \textit{Any-ttach} through two multi-tool long-horizon manipulation tasks: \textit{Sandwich Assembly} with three skills and  \textit{Cucumber Preparation} with four skills.

We demonstrate the integration of autonomous tool swapping and heterogeneous manipulation skills through two multi-tool long-horizon applications, as shown in Fig.~\ref{fig:long_horizon}(a--b).

\textit{Sandwich Assembly} combines object handling with both an active gripper and passive tools: picking and placing with the gripper, transferring an egg with a spatula, and scooping salt with a spoon.
\textit{Cucumber Preparation} represents contact-rich manipulation of object processing: peeling with a peeler, cutting with a cutter and a knife, and spearing a cucumber chunk with a fork.

These demonstrations illustrate how a single robot can coordinate tool selection, autonomous end-effector exchange, and task-specific manipulation skills across different physical interfaces.
Together with the quantitative evaluations of swapping and individual skills, they demonstrate the integration of interchangeable hardware and reusable manipulation capabilities within a unified robotic system.

\section{Conclusion and Future Work}
\label{sec:conclusion}

We presented \textit{Any-ttach}, an extensible robotic system that expands manipulation capability through autonomous end-effector swapping and reusable tool-use skills.
By integrating a passive standardized swapping mechanism, adapter generation, a matching handheld demonstration device, and a unified manipulation framework, the system supports diverse end-effectors through a simple robotic interface.
Quantitative experiments demonstrate reliable autonomous swapping, integration of 15 heterogeneous end-effector modules, and execution of tool-use skills acquired through different methods.
Two multi-tool application demonstrations further illustrate the integration of autonomous tool exchange and reusable manipulation skills in long-horizon tasks.

Beyond tool exchange, the standardized interface simplifies tool acquisition.
The robot can retrieve tools from predefined docking configurations, avoiding the need to estimate and maintain arbitrary grasp configurations during tool acquisition and use.
This shifts part of the manipulation complexity from online grasp control to the standardized mechanical interface.

One limitation of the current system is its limited ability to adapt to variations in tool–object contact during manipulation. 
Future work could introduce controller-level compliance with force/torque and contact sensing to accommodate contact variations while preserving stable tool attachment.
The same sensing could also support closed-loop monitoring of tool engagement and release during swapping.
Additionally, our current human-demonstration retargeting relies on geometric transformations and bounded trajectory smoothing.
Incorporating robot kinematic constraints, trajectory optimization, and interaction feedback could also improve cross-embodiment skill transfer.

More broadly, \textit{Any-ttach} demonstrates that manipulation capability can be expanded without concentrating all dexterity in a complex robotic end-effector.
By combining a simple standardized interface with multiple interchangeable end-effectors and reusable skills, robots can acquire new manipulation capabilities through tool adaptation and skill composition rather than increasing complexity of end-effector designs.

\section{Acknowledgment}
\label{sec:acknowledgement}
We express great thanks to all members of Duke DexLab for their support throughout this project, especially Yifei Dong for detailed proofreading feedback. 
We are grateful to Boyuan Chen and his group, Jiaxun Liu, Yinsen Jia, Yuhao Huang, Zhuoqun Chen, for their insightful discussions and feedback throughout this work, and providing constructive suggestions.
We thank Haonan Chen and Zhanpeng He for valuable project related discussions, Xunjian Yin for discussions on the VLM design, and Zhengxiao Han for helpful discussions with the VIVE tracker.
 
\bibliographystyle{IEEEtran}
\bibliography{main}

\end{document}